\documentclass{article}

\usepackage[accepted]{icml2020}

\icmltitlerunning{ConQUR: Mitigating Delusional Bias in Deep Q-learning}

\usepackage{csvsimple}
\usepackage{microtype}
\usepackage{graphicx}
\usepackage{subfigure}
\usepackage{amsmath,amssymb}
\usepackage{amsthm}
\usepackage{amsfonts}       
\usepackage{booktabs} 
\usepackage{mathtools}
\usepackage{nicefrac}
\usepackage{algorithm,algorithmic}
\usepackage{pbox}
\usepackage{enumerate}
\usepackage[small]{caption}
\usepackage{multirow}
\usepackage{booktabs}
\usepackage{gensymb}
\usepackage{bm}
\usepackage{accents}
\usepackage{tablefootnote}

\newcommand{\conqur}{\textsc{ConQUR}}

\newcommand{\1}{{\mathbf 1}}

\newcommand{\bfM}{\mathbf{M}}

\newcommand{\calS}{\mathcal{S}}
\newcommand{\calF}{\mathcal{F}}
\newcommand{\calG}{\mathcal{G}}

\newcommand{\SA}{\mathit{SA}}

\newcommand{\dqnc}[1][(\lambda)]{\mathsf{DQN}{#1}}
\newcommand{\ddqnc}[1][(\lambda)]{\mathsf{DDQN}{#1}}

\newif\ifcomments
\commentstrue
\ifcomments
  \newcommand{\colornote}[3]{{\color{#1}\bf{#2: #3}\normalfont}}
\else
  \newcommand{\colornote}[3]{}
\fi

\usepackage{hyperref}

\DeclareMathOperator*{\argmax}{arg\!max}
\DeclareMathOperator*{\vcdim}{\mathsf{VCDim}}

\newcommand{\Qhat}{\hat Q}

\newtheorem{thm}{Theorem}

\theoremstyle{definition}

\newtheoremstyle{TheoremNum}%
    {\topsep}{\topsep}
    {\itshape}
    {}
    {\bfseries}
    {.}
    { }
    {\thmname{#1}\thmnote{ \bfseries #3}}
\theoremstyle{TheoremNum}

\renewcommand{\algorithmicrequire}{\textbf{Input:}}
\renewcommand{\algorithmicensure}{\textbf{Output:}}

\begin{document}

\twocolumn[
\icmltitle{ConQUR: Mitigating Delusional Bias in Deep Q-learning}



\icmlsetsymbol{equal}{*}

\begin{icmlauthorlist}
\icmlauthor{Andy Su}{google,princeton}
\icmlauthor{Jayden Ooi}{google}
\icmlauthor{Tyler Lu}{google}
\icmlauthor{Dale Schuurmans}{google,ualberta}
\icmlauthor{Craig Boutilier}{google}
\end{icmlauthorlist}

\icmlaffiliation{princeton}{Department of Computer Science, Princeton University, New Jersey, USA}
\icmlaffiliation{google}{Google Research, Mountain View, California, USA}
\icmlaffiliation{ualberta}{Department of Computing Science, University of Alberta, Edmonton, Alberta, Canada}

\icmlcorrespondingauthor{Andy Su}{andy.2008.su@gmail.com}
\icmlcorrespondingauthor{Jayden Ooi}{jayden@alum.mit.edu}
\icmlcorrespondingauthor{Tyler Lu}{tylerlu@gmail.com}
\icmlcorrespondingauthor{Dale Schuurmans}{daes@ualberta.ca}
\icmlcorrespondingauthor{Craig Boutilier}{cebly@cs.toronto.edu}

\icmlkeywords{Machine Learning, ICML}

\vskip 0.3in
]



\printAffiliationsAndNotice{}  

\begin{abstract}
\emph{Delusional bias} is a fundamental source of error in 
approximate Q-learning.
To date, the only techniques that explicitly address
delusion require comprehensive search using tabular value estimates.
In this paper, we develop efficient methods to mitigate
delusional bias by training Q-approximators with labels 
that are ``consistent'' with the underlying greedy policy class.
We introduce a simple penalization scheme that encourages
Q-labels used \emph{across training batches}
to remain (jointly) consistent with the expressible policy class.
We also propose a search framework 
that allows multiple Q-approximators to be generated and tracked,
thus mitigating the effect of premature (implicit) policy
commitments.
Experimental results demonstrate that these methods
can improve the performance of Q-learning in a variety of Atari games,
sometimes dramatically.

\end{abstract}

\section{Introduction}
\label{sec:intro}


\emph{Q-learning } \citep{watkins:mlj92,sutton:rlbook_2ed} lies
at the heart of many of the
recent successes of deep reinforcement learning (RL)
\citep{dqn-atari,alphaGo:nature2016},
with recent advancements (e.g.,
\citet{hasselt:2010,bellemare:icml17,dueling:icml16,rainbow:2017})
helping to make it among the most widely used methods in applied RL.
Despite these successes, many properties of Q-learning are poorly understood,
and it is challenging to successfully apply deep Q-learning in practice.
Various modifications 
have been proposed
to improve convergence or approximation error
\citep{gordon:ml95,Gordon99,szepesvarismart04,melo07,maeietal:icml2010,munosetal:nips16};
but it remains difficult to reliably attain both robustness and scalability.

Recently, \citet{LuEtAl_Qdelusion:nips18} identified a source of error in
Q-learning with function approximation known as \emph{delusional bias}.
This bias arises because Q-learning updates the value of state-action pairs using estimates of (sampled)
successor-state values that can be \emph{mutually inconsistent given the policy class induced by the approximator}. 
This can result in unbounded approximation error, divergence, policy cycling,
and other undesirable behavior.
To handle delusion, 
the authors
propose a \emph{policy-consistent backup} operator
that maintains multiple Q-value estimates organized 
into
\emph{information sets}.
Each information set has its own backed-up Q-values and 
corresponding ``policy commitments'' responsible for inducing these values. 
Systematic management of these sets ensures
that only \emph{consistent} choices of maximizing actions are used to update
Q-values. All potential solutions are tracked to prevent
premature convergence on specific policy commitments. 
%
Unfortunately, the proposed algorithms use tabular representations of
Q-functions,
so while this establishes foundations
for 
delusional bias,
the function approximator is used neither for generalization nor to manage the 
size of the state/action space.
Consequently, this approach is not scalable to practical RL problems.

In this work, we develop
\conqur{} (\emph{CONsistent Q-Update Regression}),
a general framework for integrating policy-consistent backups
with regression-based function approximation for Q-learning
and for managing the search through the space of possible regressors
(i.e., information sets).  With
suitable search heuristics, the proposed framework provides
a computationally effective means for minimizing the effects of
delusional bias, while scaling to practical 
problems. 

Our main contributions are as follows.
First, we define novel augmentations of Q-regression
to increase the degree of policy consistency across training batches.
Since testing exact consistency is 
expensive,
we introduce an efficient \emph{soft-consistency penalty}
that promotes consistency of labels with earlier policy commitments.
Second, using information-set structure
\citep{LuEtAl_Qdelusion:nips18}, we define a
search space over Q-regressors to explore
multiple sets of policy commitments.
Third, we propose heuristics to guide the search,
critical given the combinatorial
nature of information sets. 
Finally, experimental results on the Atari suite \citep{bellemare:jair2013}
demonstrate that \conqur{} can add 
(sometimes dramatic) improvements to Q-learning.
These results further
show that delusion does emerge in practical applications of Q-learning.
We also
show that 
straightforward
consistency penalization on its 
own (i.e., without search) can improve both standard and double Q-learning.

\section{Background}
\label{sec:preliminaries}


\label{sec:background}

We assume a discounted, infinite horizon \emph{Markov decision process (MDP)},
$\bfM = (\calS, A, P, p_0, R, \gamma)$. 
The state space $\calS$ can reflect both
discrete and continuous features, but we take the action space $A$ to be finite
(and practically enumerable).
%
%
%
We consider \emph{Q-learning} with a function approximator
$Q_\theta$ to learn an (approximately) optimal Q-function
\citep{watkins:1989,sutton:rlbook_2ed}, drawn from
some approximation class parameterized by $\Theta$
(e.g., the weights of a neural network).
When the approximator is a deep network, we generically refer to
this as \emph{DQN}, the method at the heart of many
 RL successes \citep{dqn-atari,alphaGo:nature2016}.

For online Q-learning, at a transition $s,a,r,s'$,
the Q-update is given by:
\begin{small}
\begin{equation}
\theta \leftarrow \theta + \alpha \Big(r + \gamma\max_{a'\in A} 
    Q_\theta(s', a') - Q_\theta(s, a)\Big) \nabla_\theta Q_\theta(s, a).
    \label{eq:online_q_update}
\end{equation}
\end{small}

Batch versions of Q-learning
are similar, but fit a regressor repeatedly to
batches of training examples 
\citep{ernst_batchQ:jmlr2005,riedmiller:ecml2005}, and  
are usually more data efficient and stable than online Q-learning. 
Batch methods use a sequence of (possibly randomized)
data batches $D_1,\ldots, D_T$ to produce a sequence of regressors
$Q_{\theta_1},\ldots, Q_{\theta_T} = Q_\theta$, 
estimating
the Q-function.\footnote{We describe our approach using
batch Q-learning, but it can accommodate many variants, e.g., where 
the estimators generating max-actions and value estimates are different, as in double Q-learning
\citep{hasselt:2010,double-dqn}; indeed, we experiment with such variants.}
For each $(s,a,r,s')\in D_k$, we use a prior estimator $Q_{\theta_{k-1}}$ to bootstrap the \emph{Q-label} $q = r+\gamma\max_{a'} Q_{\theta_{k-1}}(s',a')$.
We then fit $Q_{\theta_{k}}$ to this data using a
regression procedure with a suitable loss function. 
Once trained, the (implicit) induced policy $\pi_\theta$
is the \emph{greedy policy} w.r.t.\ $Q_\theta$, i.e., $\pi_\theta(s) = \argmax_{a\in A}
Q_\theta(s,a)$.  
Let $\calF(\Theta)$ (resp., $G(\Theta)$)
be the 
class of expressible Q-functions (resp., greedy policies).

Intuitively, \emph{delusional bias} 
occurs whenever a backed-up value estimate is derived
from action choices that are not (jointly) realizable in $G(\Theta)$
\citep{LuEtAl_Qdelusion:nips18}.
Standard Q-updates back up values for each $(s,a)$ pair
by \emph{independently} choosing maximizing actions at the corresponding
next states $s'$.
However, such updates may be ``inconsistent'' under approximation:
if no policy in $G(\Theta)$ can jointly express all past action choices,
\emph{backed up values may not be realizable by any expressible policy}.
%
\citet{LuEtAl_Qdelusion:nips18} show that delusion
can manifest itself with several undesirable consequences
(e.g., divergence). Most critically,
it can prevent Q-learning from learning the optimal representable policy
in $G(\Theta)$. To address this,
they propose a non-delusional \emph{policy consistent} Q-learning 
(PCQL) algorithm that provably eliminates delusion. We refer to the
original paper for details, but review the main concepts.\footnote{While delusion may not arise in other RL approaches (e.g., policy iteration, policy gradient), our contribution focuses on mitigating delusion to derive maximum performance from widely used Q-learning methods.}

The first key concept is that of \emph{policy consistency}. 
For any 
$S\subseteq \calS$, an \emph{action assignment} $\sigma_S: S\rightarrow A$
associates an action $\sigma(s)$ with each $s\in S$. We say $\sigma$
is \emph{policy consistent} if there is a greedy policy $\pi\in G(\Theta)$
s.t.\ $\pi(s) = \sigma(s)$ for all $s\in S$.
We sometimes equate a set $\SA$ of state-action pairs with the
implied assignment $\pi(s)=a$ for all $(s,a) \in\SA$.
If $\SA$ contains multiple pairs with the same state $s$, but
different actions $a$, it is a \emph{multi-assignment} (we
use the term ``assignment'' 
when there is no risk of confusion).

In (batch) Q-learning, each new regressor 
uses training labels generated by assuming maximizing actions (under the prior
regressor) are taken at its successor states.
Let $\sigma_k$ be the collection of states and
corresponding maximizing actions used to generate labels for
regressor $Q_{\theta_k}$ (assume it is policy consistent). Suppose
we train $Q_{\theta_{k}}$ by bootstrapping on
$Q_{\theta_{k-1}}$. Now consider a training sample $(s,a,r,s')$. 
Q-learning generates label $r + \gamma\max_{a'} Q_{\theta_{k-1}}(s',a')$
for input $(s,a)$.
Notice, however, that taking action $a^*=\argmax_{a'} Q_{\theta_k}(s',a')$ at $s'$ may not be 
\emph{policy consistent} with $\sigma_k$.
Thus Q-learning will estimate a value for
$(s,a)$ assuming execution of a policy that cannot be realized
given the 
approximator. 
PCQL
prevents this by
ensuring that any assignment 
used to generate
labels is consistent with earlier assignments.
This means Q-labels will often \emph{not} be generated
using maximizing actions w.r.t.\ the prior regressor.

The second key concept is that of \emph{information sets}. One will
generally not be able to use maximizing actions to generate labels,
so tradeoffs can be made when deciding which actions to
assign to different states.
Indeed, even if it is feasible to assign a maximizing action $a$ to state $s$ early in training, say at batch $k$, since it may prevent assigning a maximizing $a'$ to $s'$ later, say batch $k+\ell$, we may want to use 
a different assignment to $s$ to give more flexibility to maximize at other states later.
PCQL does not anticipate the tradeoffs---rather
it maintains \emph{multiple information sets}, each corresponding to
a different
assignment to the states seen in the training data this far. Each
gives rise to a \emph{different Q-function estimate}, resulting in multiple
hypotheses. At the end of training, the best hypothesis is that with maximum expected value 
w.r.t.\ an initial state distribution.

PCQL provides strong convergence guarantees, but it is a tabular algorithm:
the function
approximator \emph{restricts} the policy class, but is not
used to generalize Q-values. Furthermore, its theoretical guarantees
come at a cost: it uses \emph{exact} policy consistency tests---tractable
for linear approximators, but  impractical for large problems and DQN;
and it maintains \emph{all} consistent assignments. As a result,
PCQL cannot be used for large RL problems of the
type tackled by DQN.

\section{The \conqur{} Framework}
\label{sec:conqur}

We develop the \conqur{} framework to
provide a practical approach to reducing delusion in Q-learning, specifically addressing 
the limitations of PCQL identified above. \conqur{} consists of
three main components: a practical soft-constraint
penalty that promotes policy consistency; a 
search space to structure the search over multiple regressors 
(information sets, action
assignments); and heuristic search schemes (expansion, scoring) to
find good Q-regressors.

\subsection{Preliminaries}
\label{sec:conqurprelim}

We assume a set of training data
consisting of quadruples 
$(s,a,r,s')$,
divided into (possibly non-disjoint)
\emph{batches} $D_1, \ldots, D_T$ for training.
This perspective
is quite general: online RL corresponds to $|D_i| = 1$;
offline batch training (with sufficiently exploratory data) corresponds to a
single batch (i.e., $T=1$); and online
or batch methods with replay are realized when the $D_i$ 
are generated by sampling
some data source with replacement.

For any batch $D$, let $\chi(D) = \{s' : (s,a,r,s') \in D\}$ be the set of \emph{successor states} of $D$. An \emph{action assignment} $\sigma_D$ 
for $D$ is an assignment (or multi-assignment) from $\chi(D)$ to $A$,
dictating which action $\sigma_D(s')$ is considered ``maximum'' 
when
generating a Q-label for pair $(s,a)$; i.e.,
$(s,a)$ is assigned training label 
$r + \gamma Q(s',\sigma(s'))$ rather than
$r + \gamma \max_{a'\in A} Q(s',a')$.
The set of all such assignments $\Sigma(D) = A^{\chi(D)}$
grows exponentially with $|D|$.

Given a Q-function parameterization $\Theta$,
we say $\sigma_D$ is \emph{$\Theta$-consistent (w.r.t. $D$)} if there is
some $\theta\in\Theta$ s.t.\ 
$\pi_\theta(s') = \sigma(s')$
for all $s'\in\chi(D)$.\footnote{We suppress 
mention of $D$ when
clear from context.}
This is simple policy consistency, but with notation that emphasizes
the policy class.
Let $\Sigma_\Theta(D)$ denote the set of all $\Theta$-consistent assignments 
over $D$.  The union
$\sigma_1 \cup \sigma_2$ of two assignments (over $D_1, D_2$, resp.)
is defined in the usual way.



\subsection{Consistency Penalization}
\label{sec:consistency}

Enforcing strict $\Theta$-consistency as regressors
$\theta_1, \theta_2, \ldots, \theta_T$ are generated is computationally challenging.
Suppose the assignments $\sigma_1,\ldots,\sigma_{k-1}$,
used to generate labels for $D_1, \ldots D_{k-1}$,
are jointly $\Theta$-consistent (let $\sigma_{\leq k-1}$ denote their multi-set union).
Maintaining $\Theta$-consistency when generating $\theta_{k}$
imposes two requirements. First, one must generate an assignment
$\sigma_k$ over $D_{k}$ s.t.\ $\sigma_{\leq k-1} \cup\sigma_k$ is
consistent. Even testing assignment consistency can
be problematic: for linear approximators this is a linear
feasibility program 
\citep{LuEtAl_Qdelusion:nips18} whose constraint set
grows linearly with $|D_1 \cup \ldots\cup D_{k}|$. For DNNs, this
is a complex, more expensive polynomial program. 
Second, the regressor $\theta_k$ should itself be consistent with
$\sigma_{\leq k-1} \cup\sigma_k$. This too imposes a severe burden on 
regression optimization: in the linear case, it is a constrained
least-squares problem (solvable, e.g., as a quadratic program); while with DNNs, it
can be solved, say, using a more involved projected SGD.
However, the sheer number of constraints makes this impractical.


Rather than enforcing consistency, we propose a simple, computationally
tractable scheme that ``encourages'' it: a penalty term
that can be incorporated into the regression itself.
Specifically, we add a penalty function to the usual
squared loss to encourage updates of the Q-regressors to be 
consistent with the underlying information set, i.e., the prior
action assignments used to generate its labels. 

%
When constructing $\theta_{k}$, let
$D_{\leq k} = \cup \{D_j: j \leq k\}$,
and $\sigma\in\Sigma_\Theta({D_{\leq k}})$ be the collective
assignment used to generate labels for all prior regressors (including
$\theta_k$ itself). The multiset of pairs
$B = \{(s', \sigma(s')) |  s'\in \chi(D_{\le k})\}$,  is called a
\emph{consistency buffer}.
The 
assignment need not be consistent (as we elaborate below), nor does regressor $\theta_k$ need to be consistent with $\sigma$. Instead,
we use the following
\emph{soft consistency penalty} when constructing $\theta_{k}$:
\begin{align}
C_\theta(s', a) &= \sum_{a'\in A} [Q_\theta(s', a') - Q_\theta(s', a)]_+ ,\\
C_\theta(B) &= \sum_{(s', \sigma(s'))\in B} C_\theta(s', \sigma(s')),
    \label{eq:loss_fucntion_Q_delusion}
\end{align}
where $[x]_+ = \max(0, x)$. This penalizes Q-values of actions at state $s$ that are larger than that of action $\sigma(s)$. Notice $\sigma$ is $\Theta$-consistent 
iff $\min_{\theta\in\Theta} C_\theta(B) = 0$. 
We add this penalty into our regression loss for batch $D_k$:
\begin{multline}
L_\theta(D_k, B) =
\sum_{(s,a,r,s')\in D_k} \Big[r + \gamma Q_{\theta_{k-1}}(s', \sigma(s')) - \\
Q_\theta(s, a)\Big]^2 + \lambda C_\theta(B).
\label{eq:total_loss}
\end{multline}
Here $Q_{\theta_{k}}$ is the prior estimator on which labels are bootstrapped (other regressors may be used).
%
The penalty effectively acts as a ``regularizer'' on the squared Bellman error, where $\lambda$ controls
the degree of penalization, allowing a tradeoff between Bellman error and consistency with the assignment used to generate
labels. It thus \emph{promotes} consistency without incurring the expense of 
\emph{enforcing} strict consistency.
It is straightforward to replace the classic Q-learning update \eqref{eq:online_q_update} with one
using our consistency penalty:
%
\begin{multline}
\theta_{k} \leftarrow \theta_{k-1} +  \sum_{(s, a, r, s')\in D_k} \alpha \Big[r + \gamma
    Q_{\theta_{k-1}}(s', \sigma(s')) \\
    - Q_\theta(s, a)\Big] \nabla_\theta Q_\theta(s, a)
+\alpha\lambda\nabla_\theta C_\theta(B) \Bigr\rvert_{\theta=\theta_{k-1}}.
    \label{eq:penalized_q_update}
\end{multline}
%
This scheme is quite general. First, it is agnostic as to how the prior action assignments are made
(e.g., standard maximization
w.r.t.\ the prior regressor as in DQN, Double DQN (DDQN) \citep{double-dqn}, or other variants). It can also be used in conjunction with a search through alternate assignments (see below).

%
%
%
Second, the consistency buffer $B$ may be populated in a variety of ways.
Including all max-action choices from all past training batches promotes
full consistency.
However, this may be too constraining since action choices early in training are generally informed by inaccurate value estimates. $B$ may be implemented
to focus only on more recent data (e.g., with a sliding recency window, weight decay, or subsampling); and the degree of recency bias may adapt during training (e.g., becoming more inclusive as training proceeds and the Q-function converges). Reducing the size of $B$ also has computational benefits. We discuss other ways of promoting consistency in Sec.~\ref{sec:conclusion}.

%

The proposed consistency penalty resembles the temporal-consistency loss of \citet{observe_look_further}, but our aims are very different. Their temporal consistency notion penalizes changes in a next state's Q-estimate over all actions, whereas we discourage inconsistencies in the greedy policy induced by the Q-estimator, regardless of the actual estimated values.

\subsection{The Search Space}
\label{sec:searchspace}

\begin{figure}
   \centering
   \includegraphics[width=0.40\linewidth]{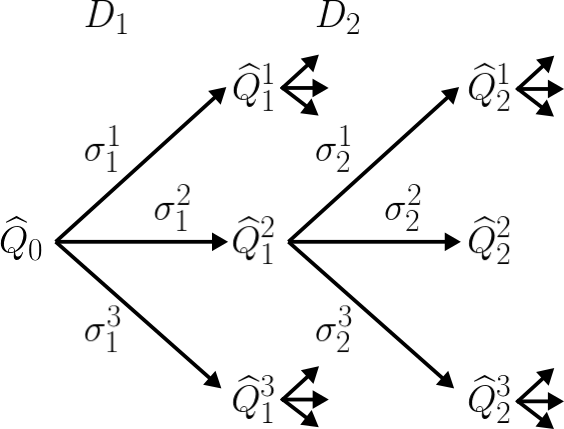}
   \caption{A generic search tree.\label{fig:searchtree}}
\end{figure}


Ensuring optimality requires that PCQL track \emph{all $\Theta$-consistent assignments}.
While the set of such assignments has polynomial size \citep{LuEtAl_Qdelusion:nips18}, it is
impractical to track in realistic problems. As such, in \conqur{} we recast information
set tracking as a \emph{search problem} and propose several strategies for managing the search
process. We begin by defining the search space and discussing its properties. We discuss
search procedures in Sec.~\ref{sec:heuristics}.

As above, assume training data is divided into batches $D_1, \ldots, D_T$ and we have some initial
Q-function estimate $\theta_0$ (for bootstrapping $D_1$'s labels). The regressor $\theta_{k}$ for $D_k$
can, in principle, be trained with labels generated by \emph{any assignment} $\sigma \in \Sigma_\Theta(D_k)$ of actions to its successor states $\chi(D_k)$, not necessarily maximizing actions w.r.t.\ $\theta_{k-1}$. Each $\sigma$
gives rise to a different updated Q-estimator $\theta_{k}$. There are several restrictions we can
place on ``reasonable'' $\sigma$-candidates: (i) $\sigma$ is $\Theta$-consistent; (ii) $\sigma$ is jointly
$\Theta$-consistent with all $\sigma_j$, for $j<k$,  used to construct the prior regressors on which
we bootstrap $\theta_{k-1}$; (iii) $\sigma$ is not \emph{dominated} by any $\sigma' \in \Sigma_\Theta(D_k)$,
where we say $\sigma'$ dominates $\sigma$ if $Q_{\theta_{k-1}}(s',\sigma'(s')) \geq Q_{\theta_{k-1}}(s',\sigma(s'))$ for all $s'\in\chi(D)$, and this inequality is strict for at least one $s'$. Conditions (i) and (ii)
are the strict consistency requirements of PCQL. We relax these below as
discussed in Sec.~\ref{sec:consistency}. Condition (iii) is inappropriate in general, since we may add additional assignments (e.g., to new data) that render all non-dominated assignments inconsistent, requiring that we revert to some dominated assignment.

This gives us a generic \emph{search space} for finding policy-consistent, delusion-free Q-function
(see Fig.~\ref{fig:searchtree}). 
Each node $n^i_k$ at depth $k$
in the search tree is associated with a regressor $\theta^i_k$ defining $Q_{\theta^i_k}$ and assignment $\sigma^i_k$ that justifies the labels used to train $\theta^i_k$ ($\sigma^i_k$ can be viewed as an information set).
The root $n_0$ is based on an initial $\theta_0$, and has an empty assignment $\sigma_0$.
%
%
Nodes at level $k$ of the tree are defined as follows. For each
node $n_{k-1}^i$ at level $k-1$---with regressor $\theta_{k-1}^i$ and
$\Theta$-consistent assignment $\sigma_{k-1}^i$---we have one child
$n_k^j$ for each $\sigma_k^j\in\Sigma_\Theta(D_k)$ such that  $\sigma_{k-1}^i \cup \sigma_k^j$
is $\Theta$-consistent. Node $n_k^j$'s assignment is $\sigma_{k-1}^i \cup \sigma_k^j$,
and its regressor $\theta_k^i$ is trained using the
data set:
$$\{(s,a) \mapsto r + \gamma Q_{\theta_{k-1}^i} (s',\sigma_k^j(s')) \,:\,
     (s,a,r,s')\in D_k\}.$$
The entire search space constructed in this fashion to a maximum depth of
$T$. 
See Appendix~\ref{app:algos}, Algorithm~\ref{alg:cellsearch} for  pseudocode of a simple depth-first recursive specification.

The exponential branching factor in this search tree would appear to make complete search intractable;
however, since we only allow $\Theta$-consistent ``collective'' assignments we can bound the size of
the tree---it is \emph{polynomial} in the VC-dimension of the approximator.
\begin{thm}
\label{thm:treesize}
The number of nodes in the search tree is no more than 
$O(nm\cdot [\binom{m}{2}n]^{\vcdim(\calG)})$ where $\vcdim(\cdot)$ is
  the VC-dimension~\citep{Vapnik1998} of a set of boolean-valued functions, and
  $\calG$ is the set of boolean functions defining all feasible greedy policies
  under $\Theta$: $\calG = \{ g_\theta(s, a, a') := \1[f_\theta(s, a) - f_\theta(s, a') > 0], \forall s,a\ne a' ~|~ \theta \in \Theta\}.$
\end{thm}
A linear approximator with a fixed set of
$d$ features induces a policy-indicator function class $\calG$
with VC-dimension $d$, making the search tree polynomial in the
size of the MDP. Similarly, a fixed ReLU DNN architecture with $W$ weights and
$L$ layers has VC-dimension of size $O(W L \log W)$ again rendering the tree
polynomially sized.

Even with this bound, navigating the search space exhaustively is generally impractical.
Instead, various search methods can be
used to explore the space, with the aim of reaching a ``high quality'' regressor
at some leaf of the tree.

\subsection{Search Heuristics}
\label{sec:heuristics}

Even with the bound in Thm.~\ref{thm:treesize}, traversing the search space exhaustively is generally impractical. Moreover, as discussed above, enforcing consistency when generating the children of a node,
and their regressors, may be intractable. Instead, various search methods can be
used to explore the space, with the aim of reaching a ``high quality'' regressor
at some (depth $T$) leaf of the tree. We outline three primary considerations in the search
process: child generation, node evaluation or scoring, and the search procedure.

{\bf Generating children.}
Given node $n^i_{k-1}$, there are, in principle, exponentially many action assignments, or children, $\Sigma_\Theta(D_k)$ (though Thm.~\ref{thm:treesize} limits this if we enforce consistency). Thus, we develop
heuristics for generating a small set of children,
driven by three primary factors.

The first factor is a preference for
generating \emph{high-value assignments}. To accurately reflect the intent of
(sampled) Bellman backups, we
prefer to assign actions to state $s'\in\chi(D_k)$ with larger predicted Q-values 
i.e., a preference for $a$ over $a'$ if ${Q}_{\theta_{k-1}^j}(s',a)>{Q}_{\theta_{k-1}^j}(s',a')$. However, since the maximizing assignment may
be $\Theta$-inconsistent (in isolation, jointly
with the parent information set, or with future assignments), candidate children
should merely have \emph{higher probability} of a high-value assignment.
Second, we need to ensure \emph{diversity} of assignments among the children.
Policy commitments at stage $k$ constrain the assignments at subsequent stages.
In many search procedures (e.g., beam search), we avoid
backtracking, so we want the stage-$k$ commitments to offer flexibility in later stages. The third factor is the degree to which we enforce consistency.

There are several ways to generate high-value assignments. We focus on one natural technique: sampling action assignments using a Boltzmann distribution.
Let $\sigma$ be the
assignment 
of some node (parent) at level $k-1$ in the tree. We generate an assignment $\sigma_k$ for $D_k$ as follows. Assume some
permutation $s_1', \ldots, s'_{|D_k|}$ of $\chi(D_k)$. For each $s'_i$ in turn, we
sample $a_i$ with probability proportional to $e^{Q_{\theta_{k-1}}(s'_i,a_i)/\tau}$. This can be
done \emph{without regard to consistency}, in which case we
use the consistency penalty
when constructing the regressor $\theta_k$ for this child to ``encourage'' consistency rather than
enforce it. If we want strict consistency, we can use rejection sampling without replacement to ensure
$a_i$ is consistent with $\sigma^j_{k-1} \cup \sigma_{\leq i-1}$ (we can also use a subset of $\sigma^j_{k-1}$ as a less restrictive consistency buffer).\footnote{Notice that at least one action for
state $s'_i$ must be consistent with any previous
(consistent) information set.} The temperature parameter $\tau$ controls the degree to which we focus on maximizing assignments versus diverse, random assignments.
%
%
While sampling gives some diversity, this procedure biases selection of high-value
actions to states 
that occur early in
the permutation. To ensure further diversity, we use a new random permutation for each child.


{\bf Scoring children.}
Once the children of some expanded node are generated,
we must assess the quality of 
each child to decide which new nodes to expand.
One possiblity is to use the
average Q-label (overall, or weighted using an initial distribution), Bellman error, or loss incurred by the regressor.
However, care
must be taken when comparing nodes
at different depths of the tree. Since deeper nodes have a greater chance to accrue rewards
or costs, simple calibration methods can be used. Alternatively, when a simulator is available, rollouts of the
induced greedy policy can be used evaluate the node quality.
However, rollouts 
incur considerable computational expense during training relative to
the more direct scoring methods.

{\bf Search Procedure.}
Given a method for generating and scoring children, 
different search procedures can be applied: best-first search, beam search, local search, etc.\ all fit very naturally within the \conqur{} framework. Moreover, hybrid strategies are possible---one we develop
below is a variant of beam search in which we generate multiple children only at certain levels of the tree,
then do ``deep dives'' using consistency-penalized Q-regression at the intervening levels. This reduces the size of the search tree considerably and, when managed properly, adds only a constant-factor (proportional to beam size) slowdown to
methods like DQN.

\subsection{An Instantiation of the \conqur{} Framework}
\label{sec:conqurdetails}

We now outline a specific instantiation of the \conqur{} framework that effectively navigates the large search spaces that arise in practical RL settings. 
We describe a heuristic, modified beam-search strategy with backtracking and priority scoring. We outline only key features (see details in Algorithm~\ref{alg:beam}, Appendix~\ref{app:algos}).

%
%
Our search process 
alternates between two phases. In an \emph{expansion phase}, parent nodes are expanded, generating one or more child nodes with assignments sampled from the Boltzmann distribution. For each child, we create target Q-labels, then optimize
its regressor using consistency-penalized Bellman error Eq.~\ref{eq:total_loss}, foregoing strict policy consistency.
In a \emph{dive phase}, each parent generates \emph{one} child, whose action assignment is given by standard
max-action selection w.r.t.\ the parent's regressor.
No diversity is considered
but we continue to use
consistency-penalized regression.

From the root, the search begins with an expansion phase to create $c$ children---$c$ is
the \emph{splitting factor}. Each child inherits its parent's consistency buffer to which we add the new assignments used for that child's Q-labels. 
To limit the tree size, we track a subset of the children (the \emph{frontier}),
selected using some scoring function. 
We select the top $\ell$-nodes for expansion, proceed to a dive phase and iterate.

We consider backtracking strategies that return to unexpanded nodes at shallower depths of the tree below.

\subsection{Related Work}
\label{sec:related}

Other work has considered multiple hypothesis tracking in RL.
One direct approach uses ensembling, 
with multiple Q-approximators updated in parallel
\citep{fausser2015neural,osband2016deep,ADQN-2018}
and  combined to reduce instability and variance.
Population-based methods,
inspired by evolutionary search, are also used.
\citet{ns-2018} combine novelty search and quality diversity
to improve hypothesis diversity and quality.
\citet{es-2018} augment an off-policy RL method
with diversified population information derived from an evolutionary algorithm. 
These techniques 
do not target a specific weaknesses of Q-learning,
such as delusion.

\section{Empirical Results}
\label{sec:empirical}

We assess the performance of
\conqur{} using the Atari test suite \citep{bellemare:jair2013}.
Since \conqur{} directly tackles delusion, any performance improvement 
over Q-learning baselines strongly suggests the presence of delusional bias
in the baselines in these domains.
 We first assess the impact of  our
 consistency penalty in isolation (without search),
treating it as
a ``regularizer'' that promotes consistency with both DQN and DDQN. We then test our modified beam search 
to assess the full power of \conqur{}. We do not directly compare \conqur{}
to policy gradient or actor-critic methods---which for some Atari games
offer state-of-the-art performance \citep{model_atari:2019,r2d2_atari:iclr2020}---because our aim with
\conqur{} is to improve the performance of (widely used) Q-learning-based algorithms.

\subsection{Consistency Penalization}
\label{subsec:exp_reg}

We first study the effects of augmenting both DQN and DDQN with soft-policy consistency in isolation. We train models using an open-source implementation
of DQN and DDQN, using default  hyperparameters \citep{tf_agents} . We refer to
the consistency-augmented algorithms as $\dqnc$ and $\ddqnc$, respectively, where $\lambda$ is the penalty weight (see Eq.~\ref{eq:total_loss}). When $\lambda = 0$, these
correspond to DQN and DDQN themselves. This policy-consistency augmentation is
lightweight and 
can be applied readily to any regression-based Q-learning method. Since we do not use search (i.e., do not track multiple hypotheses), these experiments
use a small consistency buffer drawn only from the
\emph{current} data batch by sampling from the replay buffer---this prevents getting ``trapped'' by premature policy commitments.
No diversity is used to generate action assignments---standard action maximization is used.

\begin{figure}
  \centering
  \includegraphics[width=\columnwidth]{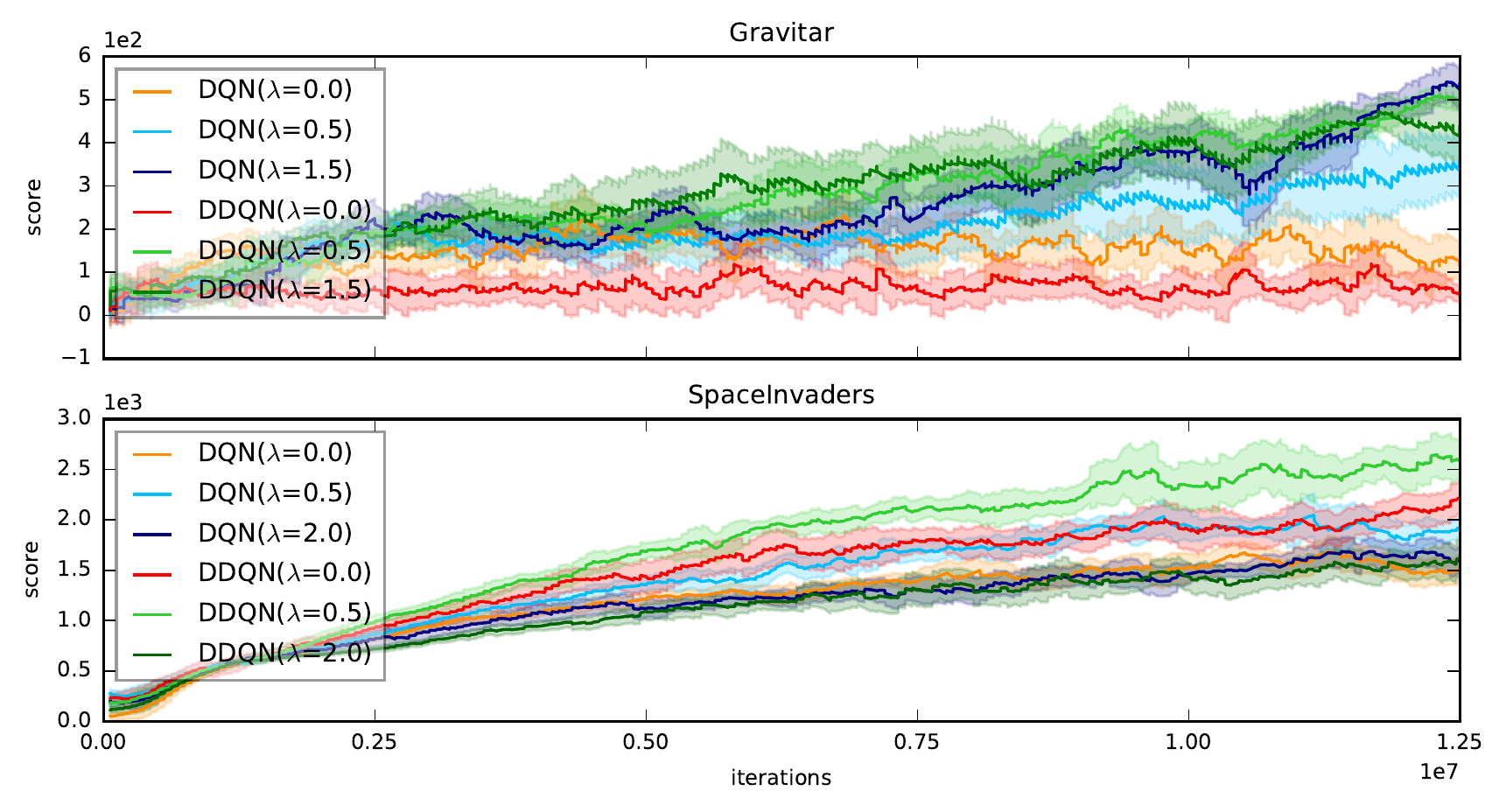}
  \caption{Varying penalization $\lambda$ (no search procedure).}
  \label{fig:e1}
\end{figure}

We evaluate $\dqnc$ and $\ddqnc$ for $\lambda \in \{0.25, 0.5 , 1, 1.5, 2 \}$ on 19 Atari games.\footnote{These 19 games were selected arbitrarily simply to test soft-consistency in isolation. See
Appendix~\ref{app:ablation_penalty} for details.} In training, $\lambda$ is initialized at $0$ and annealed to the desired value to avoid premature commitment to poor assignments.\footnote{The annealing schedule is $\lambda_t = \lambda_{\textrm{final}}  t / (t+2\times 10^6)$. Without annealing, the model tends anchor on
poorly informed assignments during early training, adversely impacting performance.}
Unsurprisingly, the best $\lambda$ tends to differ across games
depending on 
the extent of delusional bias.
Despite this, $\lambda = 0.5$ works well across all games tested. Fig.~\ref{fig:e1} illustrates the effect of increasing $\lambda$ on two games. In Gravitar, it
results in better performance in both $\dqnc[]$ and $\ddqnc[]$, while in SpaceInvaders, $\lambda = 0.5$ improves both baselines, but relative
performance degrades at $\lambda=2$.

We also compare performance on each game for each $\lambda$ value, as well as
using the best $\lambda_{\textrm{best}}$ (see Fig.~\ref{fig:all_penalty}, Table~\ref{tab:best_reg} in Appendix~\ref{app:consis_detailed}). 
$\dqnc[(\lambda_{\text{best}})]$ and $\ddqnc[(\lambda_{\text{best}})]$ outperform their ``potentially delusional'' counterparts in all but 3 and 2 games, respectively. In 9 games, both $\dqnc[(\lambda_{\text{best}})]$ and $\ddqnc[(\lambda_{\text{best}})]$ beat \emph{both} baselines. With a fixed $\lambda = 0.5$, $\dqnc$ and $\ddqnc$
each beat their respective baseline in 11 games. These results suggest that consistency penalization---independent of the general \conqur\ model---can improve the performance of DQN and DDQN by addressing delusional bias.
Moreover, promoting policy consistency appears to have a different effect on learning
than double Q-learning, which addresses maximization bias. Indeed, consistency penalization, when applied to $\dqnc[]$, achieves greater gains than $\ddqnc[]$ in 15 games. Finally, in 9 games $\ddqnc$ improves unaugmented $\dqnc$.
Further experiment details and results can be found in Appendix~\ref{app:ablation_penalty}.





%
%

\subsection{Full \conqur{}}

\label{subsec:exp_conqur}
\begin{figure}[t]
\centering
\begin{minipage}{.45\columnwidth}
  \centering
  \includegraphics[width=\columnwidth]{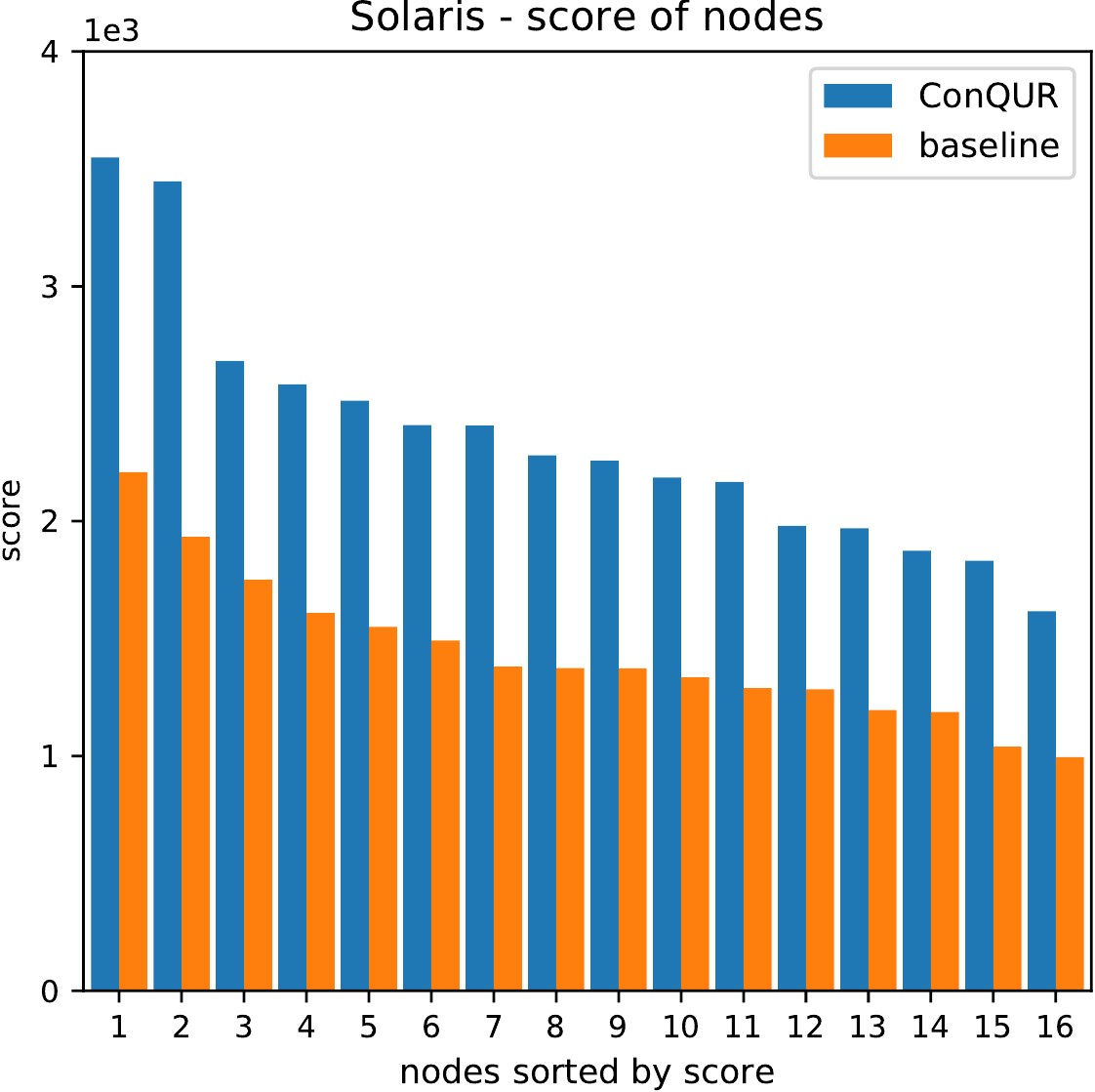}
  \captionof{figure}{Policy value (game scores) of nodes, sorted.}
  \label{fig:cells_bar}
\end{minipage}
\;
\begin{minipage}{.45\columnwidth}
  \centering
  \includegraphics[width=\columnwidth, height=3.5cm]{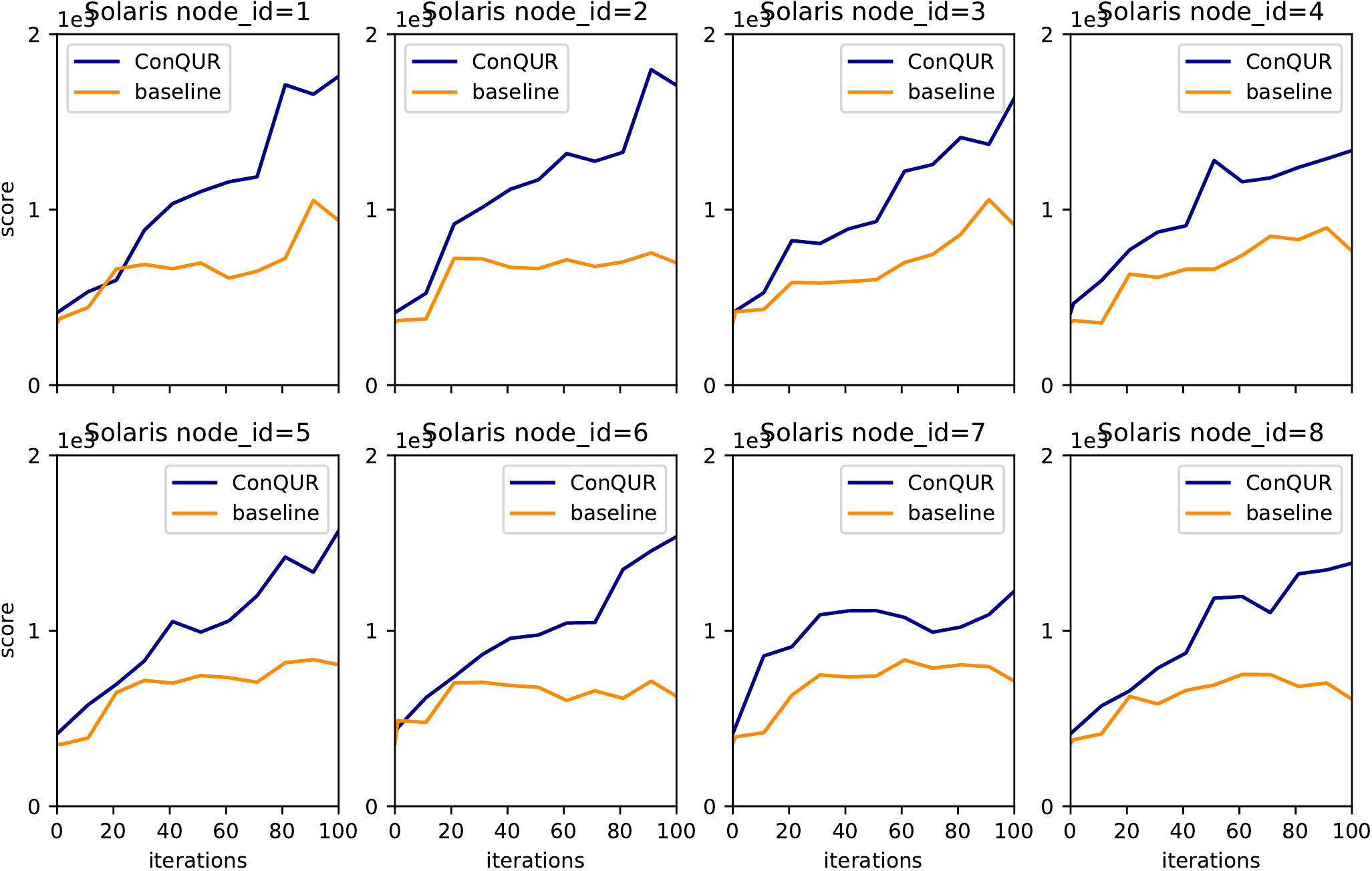}
  \captionof{figure}{Training curves of 8 sorted nodes.}
  \label{fig:cells}
\end{minipage}
\end{figure}
\begin{figure}[t]
\centering
   \includegraphics[width=\columnwidth]{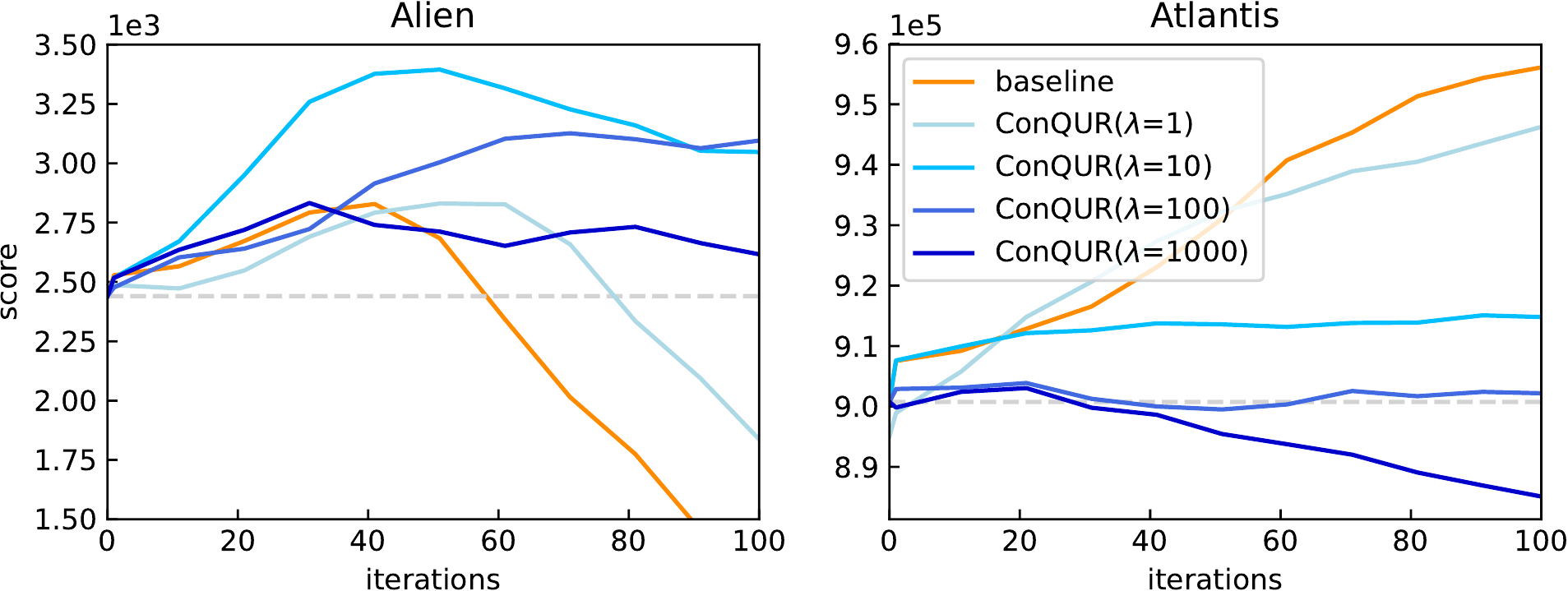}
  \captionof{figure}{Performance effects of varying $\lambda$.}
  \label{fig:conqur_reg}
\end{figure}

\begin{figure}[htb]
\begin{center}
    \centerline{\includegraphics[width=0.98\columnwidth, scale=0.9]{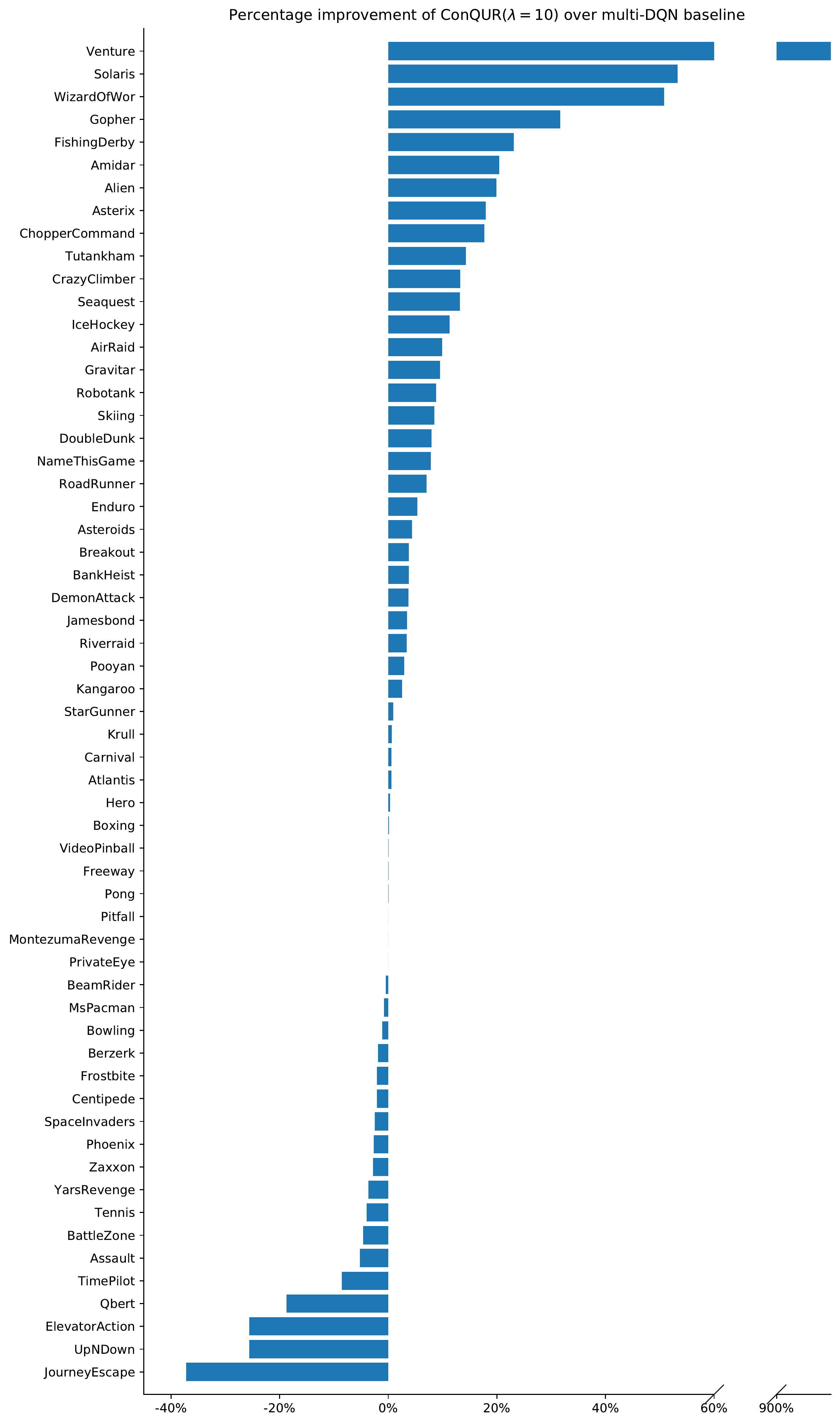}}
\caption{Improvements of ConQUR($\lambda=10$) over multi-DQN baseline on all 59 games. A frontier $F=16$ nodes was used.}
\label{fig:winning_margin}
\end{center}
\end{figure}
We test the full \conqur\ framework using our modified beam search (Sec.~\ref{sec:conqurdetails}) on the full suite of 59 Atari games.
Rather than training a full Q-network using \conqur{}, 
we leverage
pre-trained networks from the Dopamine package \cite{castro18dopamine},\footnote{See {\small {\tt https://github.com/google/dopamine}}} and use \conqur{} to learn
final layer weights, i.e., a new ``linear approximator'' w.r.t.\ the learned
feature representation.
We do this for two reasons. First, this allows us to test whether
delusional bias occurs in practice. By freezing the learned representation, any
improvements offered by \conqur{} when learning a
linear Q-function over those same features provides direct
evidence that (a) delusion is present in the original trained baselines,
and (b) \conqur{} does in fact mitigate its impact (without relying on
novel feature discovery).
Second, from a practical point of view, this ``linear tuning'' approach offers a relatively inexpensive way to apply our methodology in practice. By bootstrapping
a model trained in standard fashion and extracting performance gains with
a relatively small amount of additional training
(e.g., linear tuning requires many fewer training samples, as our results show), we can offset
the cost of the \conqur{} search process itself.

We use DQN-networks with the same architecture as in~\citet{dqn-atari}, trained on 200M frames as our baseline.
We use \conqur\ to retrain only the last (fully connected) layer (freezing other layers), which can be viewed as a linear Q-approximator over the features learned by the CNN. We train Q-regressors in \conqur\ using \emph{only 4M additional frames}.\footnote{This reduces computational/memory footprint
of our experiments, and suffices since we re-train a simpler
approximator. Nothing in the framework \emph{requires} this reduced training data.}
We use a splitting factor of $c=4$ and
frontier size 16. The dive phase is always of length nine (i.e., nine batches of data), giving an expansion phase every ten iterations. Regressors are trained using soft-policy consistency (Eq.~\ref{eq:total_loss}), with the consistency buffer comprising \emph{all} prior action assignments.
We run \conqur{} with $\lambda \in \{1,10\}$ and select the best performing policy. 
We use larger $\lambda$ values than in Sec.~\ref{subsec:exp_reg} since
full \conqur{} maintains multiple Q-regressors and can ``discard'' poor
performers. This allows more aggressive consistency enforcement---in the extreme, with exhaustive search and $\lambda\to \infty$, \conqur{} behaves like PCQL, finding a near-optimal greedy policy. See
Appendix~\ref{app:full_atari} for further details (e.g., hyperparameters) and  results. 

%

We first test two approaches to scoring nodes: (i) policy evaluation using rollouts; and (ii) scoring using the loss
function (Bellman error with soft consistency). Results on a small selection of games are shown in Table~\ref{tab:scoringfn}. While rollouts,  unsurprisingly, tend to 
induce better-performing policies, consistent-Bellman scoring is competitive. Since the latter much less computationally intense, and does not require a
simulator (or otherwise sampling the environment), we use it throughout our remaining experiments.

We next compare \conqur{} with the value of the pre-trained DQN. We also evaluate a ``multi-DQN''
baseline that trains multiple DQNs independently, warm-starting from the same
pre-trained DQN. It uses the same number of frontier nodes as \conqur, and is
trained identically to \conqur, but uses direct Bellman error (no consistency penalty). This gives DQN the same advantage of multiple-hypothesis tracking as \conqur\ (without its policy consistency).

\begin{table}[t]
\begin{center}
\begin{small}
\begin{tabular}{|l|r|r|}
\hline 
 & \textbf{Rollouts} & \textbf{Bellman + Consistency Penalty} \\ \hline
 BattleZone & 33796.30 & 32618.18 \\
 BeamRider & 9914.00 & 10341.20 \\
 Boxing & 83.34 & 83.03 \\
 Breakout & 379.21 & 393.00 \\
 MsPacman & 5947.78 & 5365.06 \\
 Seaquest & 2848.04 & 3000.78 \\
 SpaceInvader & 3442.31 & 3632.25 \\
 StarGunner & 55800.00 & 56695.35 \\
 Zaxxon & 11064.00 & 10473.08 \\ \hline
\end{tabular}
\end{small}
\end{center}
\caption{Results of \conqur{} with 8 nodes (split 2) on 9 games: comparing
loss-based node scoring with scoring using rollouts.}
\label{tab:scoringfn}
\end{table}

We test on 59 games.  \conqur{} with frontier size 16 and expansion factor 4 and splitting factor 4 (16-4-4) with backtracking (as described in the Appendix~\ref{app:full_atari}) results in significant improvements over the pre-trained DQN, with an average score improvement of 189\%. The only games without improvement are Montezuma's Revenge, Tennis, Freeway, Pong, PrivateEye and BankHeist. This demonstrates that, \emph{even when simply
retraining the last layer of a highly tuned DQN network}, removing delusional bias 
frequently improves policy performance significantly. \conqur{} exploits the reduced
parameterization to obtain these gains with only 4M frames of training data. A half-dozen games have outsized improvements over pre-trained DQN, including Venture (35 times greater value), ElevatorAction (23 times), Tutankham (5 times) and Solaris (5 times).\footnote{This may be in part, but not fully, due to the sticky-action training of the pre-trained model.}

We found that $\lambda=10$ provided the best performance across all games. Fig.~\ref{fig:winning_margin} shows the  percentage improvement of \conqur{($\lambda=10$)} over the multi-DQN baseline for all 59 games. The improvement is defined as $(s_C - s_B)/|s_B|$ where $s_C$ and $s_B$ are the average scores (over 5 runs) of the policy generated by \conqur{} and that by the multi-DQN baseline (16 nodes), respectively. Compared to this stronger baseline, \conqur{} wins by 
a margin of at least 10\% in 16 games, while 19 games see improvements of 1--10\%,
16 games show little effect ($\pm 1$\%) and 8 games show a decline of greater than 1\%. 
Tables of complete results and figures of training curves (all games) appears in Appendix~\ref{subapp:conqur_results}, Table~\ref{tab:allnoop} and Fig.~\ref{fig:allgames_allreg}.


Figs.~\ref{fig:cells_bar} and~\ref{fig:cells} (smoothed, best frontier node) show node policy values and training curves, respectively, for Solaris. When examining nodes ranked by their policy value
(Fig.~\ref{fig:cells_bar}), we see that nodes of any given rank generated by \conqur{} dominate
their by multi-DQN (baseline) counterparts: the three highest-ranked nodes exceed
their baseline counterparts by 18\%, 13\% and 15\%, respectively, while the remaining nodes show improvements of roughly 11--12\%. Fig.~\ref{fig:conqur_reg} (smoothed, best frontier node) shows the effect of varying $\lambda$. In Alien, increasing $\lambda$ from 1 to 10 improves performance, but 
performance starts to decline for higher $\lambda$ (we tested both 100 and 1000). This is similar to patterns observed in Sec.~\ref{subsec:exp_reg} and represents a trade-off between emphasizing consistency and not over-committing to action assignments. In Atlantis, stronger penalization tends to degrade performance. In fact, the stronger the penalization, the worse the performance. 

\section{Concluding Remarks}
\label{sec:conclusion}

We have introduced \conqur{}, a framework for mitigating delusional bias in various
forms of Q-learning that relaxes
some of the strict assumptions of exact delusion-free algorithms
like PCQL to ensure scalability. Its main
components are a search procedure used to
maintain diverse, promising Q-regressors (and
corresponding information sets); and a consistency penalty that encourages ``maximizing'' actions to be consistent with the approximator class. \conqur{} embodies elements of both value-based and policy-based RL: it can be viewed as using partial policy constraints to bias the Q- value estimator, and as a means of using candidate value functions to bias the search through policy space. Empirically, we find that \conqur{} can improve the quality of existing approximators by removing delusional bias. Moreover, the consistency penalty applied on its own,
in either DQN or DDQN, can improve policy quality.

There are many directions for future research. Other methods for nudging regressors to be policy-consistent include exact consistency (i.e., constrained regression), other regularization schemes that push the regressor to fall within the information set, etc. Further exploration of search,
child-generation, and node-scoring strategies should be examined within \conqur{}. Our (full) experiments should also be extended beyond those that warm-start from a DQN model.
We believe our methods can be extended to both
continuous actions and soft max-action policies. We are also interested
in the potential connection between maintaining multiple ``hypotheses'' (i.e., Q-regressors) and notions in distributional RL
\cite{bellemare:icml17}.


\bibliographystyle{icml2020}

\bibliography{long,standard}

\clearpage
\appendix

\section{An Example of Delusional Bias}
\label{app:delusionexample}

\begin{figure}[t]
\begin{center}
\centerline{\includegraphics[width=0.9\columnwidth]{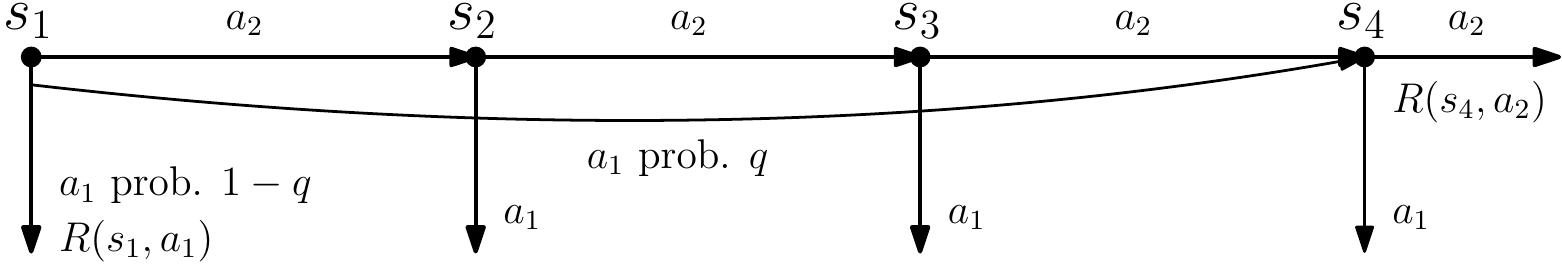}}
    \caption{A simple MDP \citep{LuEtAl_Qdelusion:nips18}.
}
\label{fig:4chain}
\end{center}
\end{figure}

We describe an example, taken directly from \citep{LuEtAl_Qdelusion:nips18},
to show concretely how delusional bias causes problems for Q-learning with
function approximation.
The MDP in Fig.~\ref{fig:4chain} illustrates the phenomenon:
\citet{LuEtAl_Qdelusion:nips18} use a linear approximator
over a specific set of features in this MDP to show that:
\begin{enumerate}[(a)]
    \item No $\pi\in G(\Theta)$ can express the optimal (unconstrained) policy (which requires
     taking $a_2$ at each state);
     \item The optimal \emph{feasible} policy in $G(\Theta)$ takes $a_1$ at $s_1$ and $a_2$ at $s_4$ (achieving a value of $0.5$).
     \item Online Q-learning (Eq.~\ref{eq:online_q_update}) with data generated using an $\varepsilon$-greedy behavior policy must converge to a fixed point (under a range of rewards and discounts) corresponding to
      a ``compromise'' admissible policy which takes $a_1$ at both $s_1$ and $s_4$ (value of $0.3$).
\end{enumerate}
Q-learning fails to find a reasonable fixed-point 
because of delusion. Consider the backups at $(s_2, a_2)$ and $(s_3, a_2)$.
Suppose $\hat{\theta}$ assigns a ``high'' value to $(s_3,a_2)$, so that $Q_{\hat{\theta}}(s_3,a_2) > Q_{\hat{\theta}}(s_3,a_1)$ as required by $\pi_{\theta^{\ast}}$. They show that
any such $\hat{\theta}$ also accords a ``high'' value to $(s_2, a_2)$.
But $Q_{\hat{\theta}}(s_2,a_2) > Q_{\hat{\theta}}(s_2,a_1)$ is
inconsistent the first requirement. As such,
any update that makes the Q-value of $(s_2,a_2)$ higher
\emph{undercuts the justification} for it to be higher (i.e., makes the ``max''
value of its successor state $(s_3,a_2)$ lower).
This occurs not due to approximation
error, but the inability of Q-learning to find the value of the
optimal \emph{representable} policy.

\section{Algorithms}
\label{app:algos}
The pseudocode of (depth-first) version of the \conqur{} search framework is listed in Algorithm~\ref{alg:cellsearch}.
\begin{algorithm}[bt]
 \caption{\conqur{} \textsc{Search} (Generic, depth-first)}
 \label{alg:cellsearch}
  \begin{algorithmic}[1]
    \REQUIRE Data sets $D_k, D_{k+1}, \ldots D_{T}$; regressor $\Qhat_{k-1}$; and assignment
             $\sigma$ over $D_{\leq k-1}= \cup_{1\leq j \leq k-1} D_{j}$ reflecting
             prior data; policy class $\Theta$.\\
    \STATE Let $\Sigma_{\Theta,\sigma} = \{\sigma_k\in\Sigma_\Theta(D_j) :
        \sigma_k \cup \sigma \textrm{ is consistent}\}$ \label{line:consistentsigma}
    \FOR{all $\sigma_k^j \in \Sigma_{\Theta,\sigma}$}
      \STATE Training set $S\leftarrow \{\}$
      \FOR{all $(s,a,r,s')\in D_k$}
        \STATE $q \leftarrow r + \gamma \Qhat_{k-1}(s',\sigma_k^j(s'))$
        \STATE $S\leftarrow S\cup\{((s,a),q)\}$
      \ENDFOR
      \STATE Train $\Qhat^j_{k}$ using training set $S$ \label{line:training}
      \IF{$k=T$}
        \STATE Return $\Qhat^j_{k}$  // terminate \label{line:terminate}
      \ELSE
        \STATE Return \textsc{Search}($D_{k+1}, \ldots D_{T}; \, \Qhat^j_{k} ; \, \sigma_k^j \cup
              \sigma ; \, \Theta$) // recurse \label{line:recurse}
      \ENDIF
    \ENDFOR
  \end{algorithmic}
\end{algorithm}
As discussed in Sec.~\ref{sec:conqurdetails}, a more specific instantiation of the \conqur{} algorithm is listed in Algorithm~\ref{alg:beam}.
\begin{algorithm*}[htb]
\caption{Modified Beam Search Instantiation of \conqur{} Algorithm}
\label{alg:beam}
\begin{algorithmic}[1]
  \REQUIRE Search control parameters: $m$, $\ell$, $c$, $d$, $T$
  \STATE {\rm Maintain} list of data batches $D_1,...,D_k$, initialized empty
  \STATE {\rm Maintain} candidate pool $P$ of at most $m$ nodes, initialized $P=\{n_0\}$
  \STATE {\rm Maintain} frontier list $F$ of $\ell^c$ nodes
  \STATE {\rm Maintain} for each node $n^i_k$ a regressor $\theta^i_k$ and an ancestor assignment $\sigma^i_k$
\vskip1ex
  \FOR{ each search level $k\leq T$ }
    \STATE Find top scoring node $n^1\in P$
    \STATE Use $\varepsilon$-greedy policy extracted from $Q_{\theta^1}$ to collect next data batch $D_k$
\vskip1ex
    \IF{ $k$ is an expansion level }
      \STATE Select top $\ell$ scoring nodes $n^1,...,n^\ell\in P$
      \FOR{ each selected node $n^i$ }
        \STATE Generate $c$ children $n^{i,1},...,n^{i,c}$ using Boltzmann sampling on $D_k$ with $Q_{\theta^i}$
        \FOR{ each child $n^{i,j}$ }
          \STATE Let assignment history $\sigma^{i,j}$ be $\sigma^i\cup\{\textit{new assignment}\}$
          \STATE Determine regressor $\theta^{i,j}$ by applying update \eqref{eq:penalized_q_update} from $\theta^i$
        \ENDFOR
        \STATE Score and add child nodes to the candidate pool $P$
        \STATE Assign frontier nodes to set of child nodes, $F=\{n^{i,j}\}$
        \IF{ $|P|>m$ }
          \STATE evict bottom scoring nodes, keeping top $m$ in $P$
        \ENDIF
      \ENDFOR
    \ENDIF
\vskip1ex
    \IF{ $k$ is a refinement ("dive") level }
      \FOR{ each frontier node $n^{i,j}\in F$ }
        \STATE Update regressor $\theta^{i,j}$ by applying update \eqref{eq:penalized_q_update} to $\theta^{i,j}$
      \ENDFOR
    \ENDIF
\vskip1ex
    \STATE Run $d$ "dive" levels after each expansion level
\vskip1ex
  \ENDFOR
\end{algorithmic}
\end{algorithm*}

\section{Additional Detail: Effects of Consistency Penalization}
\label{app:ablation_penalty}

\subsection{Delusional bias in DQN and DDQN}
Both DQN and DDQN uses a delayed version of the $Q$-network $Q_{\theta^-}(s', a')$ for label generation, but in a different way. In DQN, $Q_{\theta^-}(s', a')$ is used for both value estimate and action assignment $\sigma_{\text{DQN}}(s') = \argmax_{a'} Q_{\theta_{k}}(s', a')$, whereas in DDQN, $Q_{\theta^-}(s', a')$ is used only for value estimate and the action assignment is computed from the current network $\sigma_{\text{DDQN}}(s') = \argmax_{a'} Q_{\theta_{k}}(s', a')$.

With respect to delusional bias, action assignment of DQN is consistent for all batches after the latest network weight transfer, as $\sigma_{\text{DQN}}(s')$ is computed from the same $Q_{\theta^-}(s', a')$ network. DDQN, on the other hand, could have very inconsistent assignments, since the action is computed from the current network that is being updated at every step.

\subsection{Training Methodology and Hyperparameters}
We implement consistency penalty on top of the DQN and DDQN algorithm by modifying the open-source TF-Agents library \citep{tf_agents}. In particular, we modify existing \texttt{DqnAgent} and \texttt{DdqnAgent} by adding a consistency penalty term to the original TD loss.

We use TF-Agents implementation of DQN training on Atari with the default hyperparameters, which are mostly the same as that used in the original DQN paper \citep{dqn-atari}. For conveniece to the reader, some important hyperparameters are listed in Table~\ref{tab:hps_cp}. The reward is clipped between $[-1, 1]$ following the original DQN.

\begin{table*}[t]
\begin{center}
\begin{tabular}{|l|c|}
\hline 
 Hyper-parameter & Value \\
\hline
 Mini-batch size & 32\\
 Replay buffer capacity & 1 million transitions\\
 Discount factor $\gamma$ & 0.99\\
 Optimizer & RMSProp \\
 Learning rate & 0.00025\\
 Convolution channel & $32, 64, 64$ \\
 Convolution filter size & $(8 \times 8), (4 \times 4), (3 \times 3)$ \\
 Convolution stride & 4, 2, 1\\ 
 Fully-connected hidden units & 512\\
 Train exploration $\varepsilon_{\text{train}}$ & 0.01\\
 Eval exploration $\varepsilon_{\text{eval}}$ & 0.001 \\
\hline
\end{tabular}
\end{center}
\caption{Hyperparameters for training DQN and DDQN with consistency penalty on Atari.}
\label{tab:hps_cp}
\end{table*}

\subsection{Evaluation Methodology}
We empirically evaluate our modified DQN and DDQN agents trained with consistency penalty on 15 Atari games. Evaluation is run using the training and evaluation framework for Atari provided in TF-Agents without any modifications.

\subsection{Detailed Results}
\label{app:consis_detailed}
\begin{figure*}[t]
\begin{center}
    \centerline{\includegraphics[width=1.4\columnwidth]{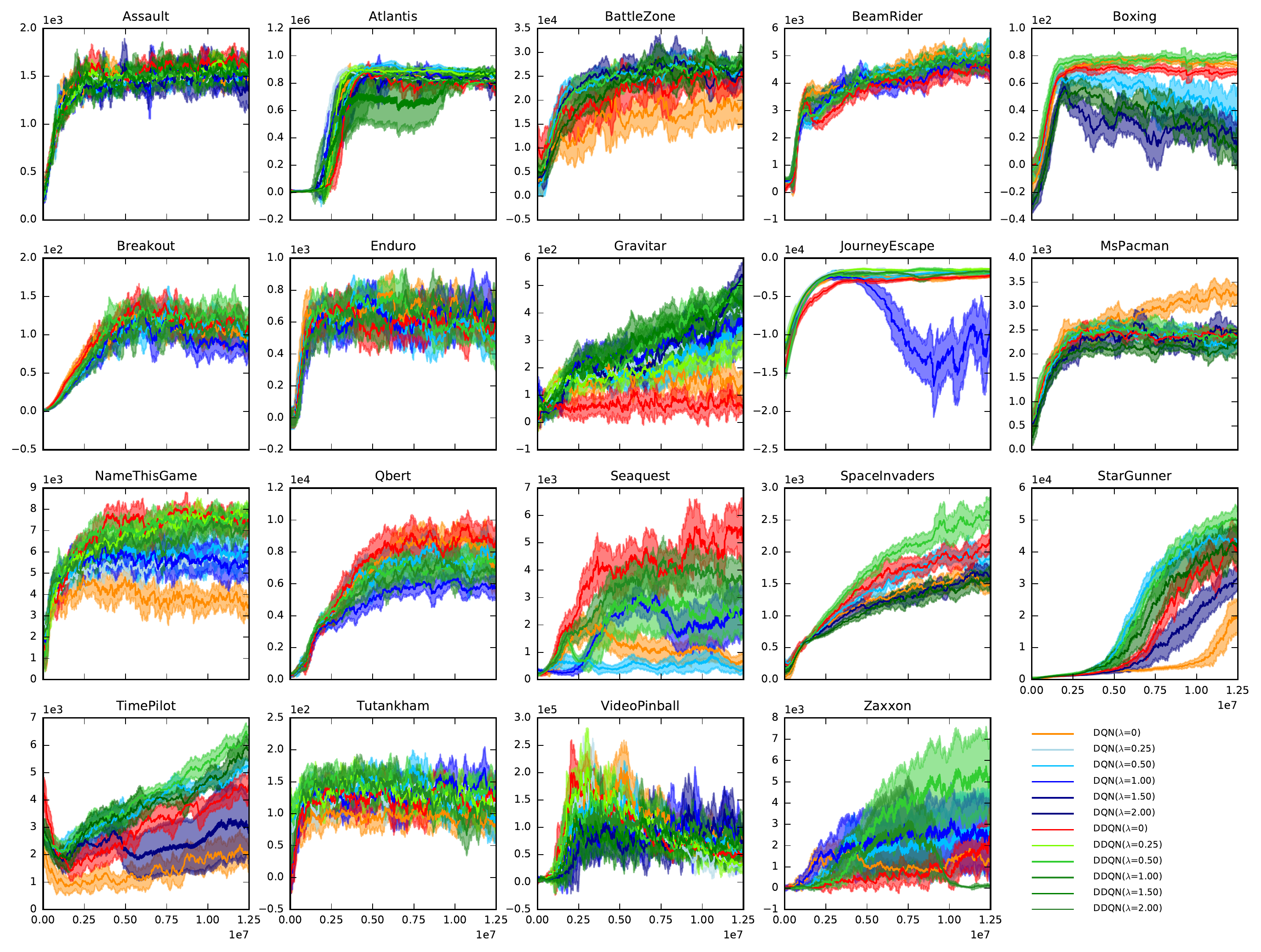}}
\caption{DQN and DDQN training curves for different penalty constant $\lambda$.}
\label{fig:all_penalty}
\end{center}
\end{figure*}

\begin{figure*}[htb]
\begin{center}
    \centerline{\includegraphics[width=1.4\columnwidth]{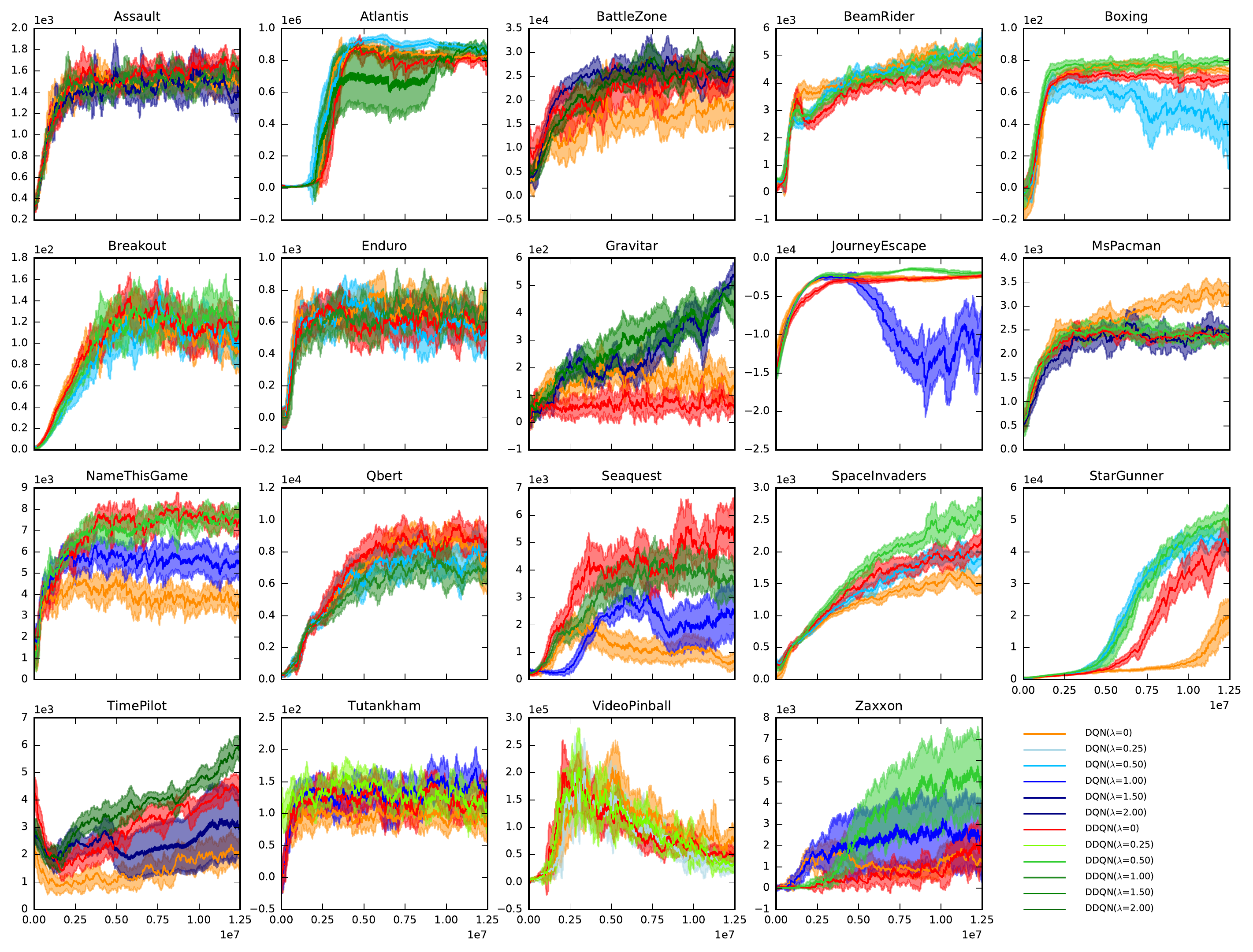}}
\caption{DQN and DDQN training curves for the respective best $\lambda$ and baseline.}
\label{fig:best_penalty}
\end{center}
\end{figure*}

\begin{figure*}[htb]
\begin{center}
    \centerline{\includegraphics[width=1.4\columnwidth]{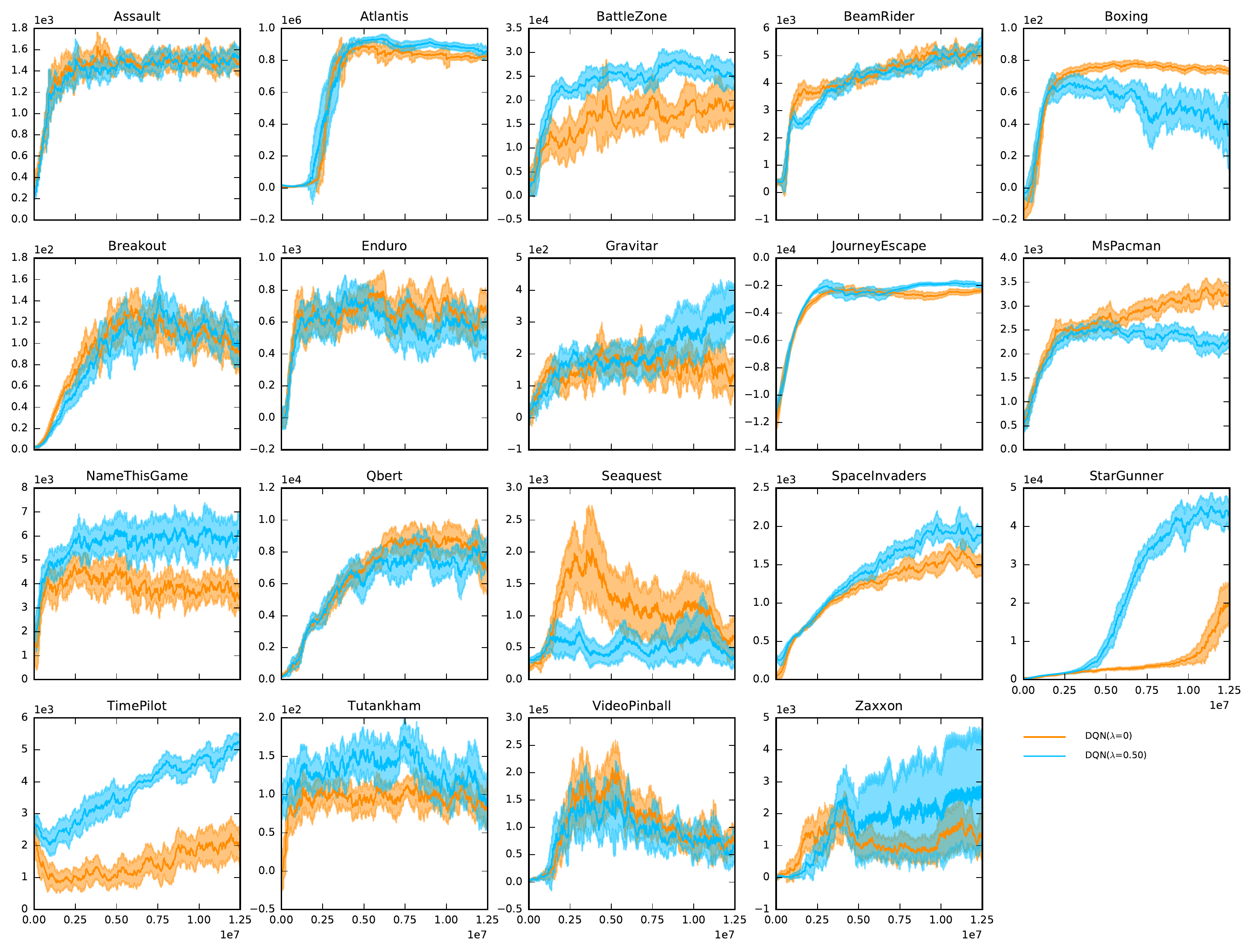}}
    \centerline{\includegraphics[width=1.4\columnwidth]{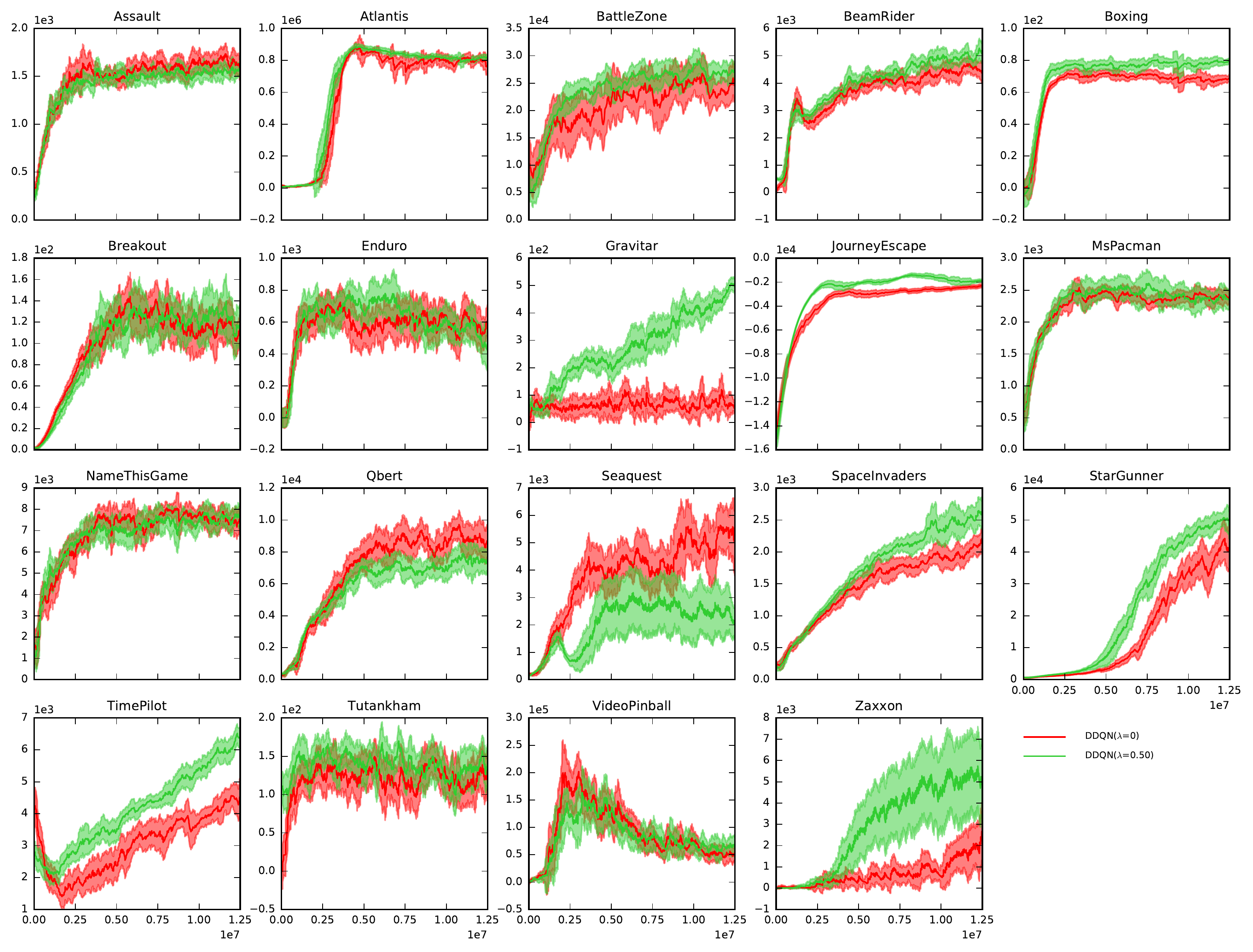}}
\caption{DQN and DDQN training curves for $\lambda = 0.5$ and the baseline.}
\label{fig:0.5_penalty}
\end{center}
\end{figure*}

Fig.~\ref{fig:all_penalty} shows the effects of varying $\lambda$ on both DQN and DDQN. Table~\ref{tab:best_reg} summarizes the best penalties for each game and their corresponding scores. Fig.~\ref{fig:best_penalty} shows the training curves of the best penalization constants.
Finally, Fig.~\ref{fig:0.5_penalty} shows the training curves for a fixed penalization of $\lambda=0.5$.
The datapoints in each plot of the aforementioned figures are obtained by averaging over window size of 30 steps, and within each window, we take the largest policy value (and over $\approx$2--5 multiple runs). This is done to reduce visual clutter.
\begin{table*}
\centering
\begin{tabular}{|l|r|c|r|r|c|r|}\hline
 & $\dqnc[]$ & $\lambda_{\text{best}}$ & $\dqnc[(\lambda_{\text{best}})]$ & $\ddqnc[]$  & $\lambda'_{\text{best}}$ & $\ddqnc[(\lambda'_{\text{best}})]$ \\
\hline
Assault & 2546.56 & 1.5 & \bf{3451.07} & 2770.26 & 1 & 2985.74 \\
Atlantis & 995460.00 & 0.5 & \bf{1003600.00} & 940080.00 & 1.5 & 999680.00 \\
BattleZone & \bf{67500.00} & 2 & 55257.14 & 47025.00 & 2 & 48947.37 \\
BeamRider & 7124.90 & 0.5 & \bf{7216.14} & 5926.59 & 0.5 & 6784.97 \\
Boxing & 86.76 & 0.5 & 90.01 & 82.80 & 0.5 & \bf{91.29} \\
Breakout & 220.00 & 0.5 & 219.15 & 214.25 & 0.5 & \bf{242.73} \\
Enduro & 1206.22 & 0.5 & \bf{1430.38} & 1160.44 & 1 & 1287.50 \\
Gravitar & 475.00 & 1.5 & \bf{685.76} & 462.94 & 1.5 & 679.33 \\
JourneyEscape & -1020.59 & 0.25 & -696.47 & -794.71 & 1 & \bf{-692.35} \\
MsPacman & \bf{4104.59} & 2 & 4072.12 & 3859.64 & 0.5 & 4008.91 \\
NameThisGame & 7230.71 & 1 & 9013.48 & 9618.18 & 0.5 & \bf{10210.00} \\
Qbert & 13270.64 & 0.5 & \bf{14111.11} & 13388.92 & 1 & 12884.74 \\
Seaquest & 5849.80 & 1 & 6123.72 & \bf{12062.50} & 1 & 7969.77 \\
SpaceInvaders & 2389.22 & 0.5 & 2707.83 & 3007.72 & 0.5 & \bf{4080.57} \\
StarGunner & 40393.75 & 0.5 & 55931.71 & 55957.89 & 0.5 & \bf{60035.90} \\
TimePilot & 4205.83 & 2 & 7612.50 & 6654.44 & 2 & \bf{7964.10} \\
Tutankham & 222.76 & 1 & \bf{265.86} & 243.20 & 0.25 & 247.17 \\
VideoPinball & \bf{569502.19} & 0.25 & 552456.00 & 509373.50 & 0.25 & 562961.50 \\
Zaxxon & 5533.33 & 1 & \bf{10520.00} & 7786.00 & 0.5 & 10333.33 \\
\hline
\end{tabular}
\caption{Consistency penalty ablation results on best penalty constants for DQN and DDQN.}
\label{tab:best_reg}
\end{table*}

\section{Additional Detail: \conqur{} Results}
\label{app:full_atari}
Our results use a frontier queue of size ($F$) 16 (these are the top scoring leaf nodes which receive gradient updates and rollout evaluations during training).
To generate training batches, we select the best node's regressor according to our scoring function, from which we generate training samples (transitions) using $\varepsilon$-greedy.
Results are reported in Table~\ref{tab:allnoop}, and training curves in Fig.~\ref{fig:allgames_allreg}.
We used Bellman error plus consistency penalty as our scoring function. During the training process, we also calibrated the scoring to account for the depth difference between the leaf nodes at the frontier versus the leaf nodes in the candidate pool. We calibrated by taking the mean of the difference between scores of the current nodes in the frontier with their parents. We scaled this difference by multiplying with a constant of 2.5.

In our implementation, we initialized our Q-network with a pre-trained DQN. We start with the expansion phase. During this phase, each parent node splits into $l$ children nodes and the Q-labels are generated using action assignments from the Boltzmann sampling procedure, in order to create high quality and diversified children. We start the dive phase until the number of children generated is at least $F$. In particular, with $F = 16$ configuration, we performed the expansion phase at the zero-th and first iterations, and then at every tenth iteration starting at iteration 10, then at 20, and so on until ending at iteration 90.
All other iterations execute the ``dive'' phase. For every fifth iteration, Q-labels are generated from action assignments sampled according to the Boltzmann distribution. For all other iterations, Q-labels are generated in the same fashion as the standard Q-learning (taking the max Q-value). The generated Q-labels along with the consistency penalty are then converted into gradient updates that applies to one or more generated children nodes.

\subsection{Training Methodology and Hyperparameters}
Each iteration consists of 10k transitions sampled from the environment. Our entire training process has 100 iterations which
consumes 1M transitions or 4M frames. We used RMSProp as the optimizer with a learning rate of 
$2.5\times 10^{-6}$. One training iteration has 2.5k gradient updates and we used a batch size of 32. We replace
the target network with the online network every fifth iteration and reward is clipped between $[-1, 1]$. We use a discount value of $\gamma =0.99$ and $\varepsilon$-greedy with $\varepsilon=0.01$ for exploration. Details of hyper-parameter settings can be found in Table~\ref{tab:hps}, \ref{tab:hps_nodes}.

\subsection{Evaluation Methodology}
We empirically evaluate our algorithms on 59 Atari games  \citep{bellemare:jair2013}, and followed the evaluation procedure as in \citet{double-dqn}. We evaluate our agents on every 10-th iteration (and also the initial and first iteration) by suspending our training process. We evaluate on 500k frames, and we cap the length of the episodes for 108k frames. We used $\varepsilon$-greedy as the evaluation policy with $\varepsilon =0.001$. We evaluated our algorithm under the \emph{no-op starts} regime---in this setting, we insert a random number of ``do-nothing'' (or \emph{no-op}) actions (up to 30) at the beginning of each episode.

\subsection{Detailed Results}
\label{subapp:conqur_results}
\begin{figure*}[htb]
\begin{center}
    \centerline{\includegraphics[width=0.90\textwidth]{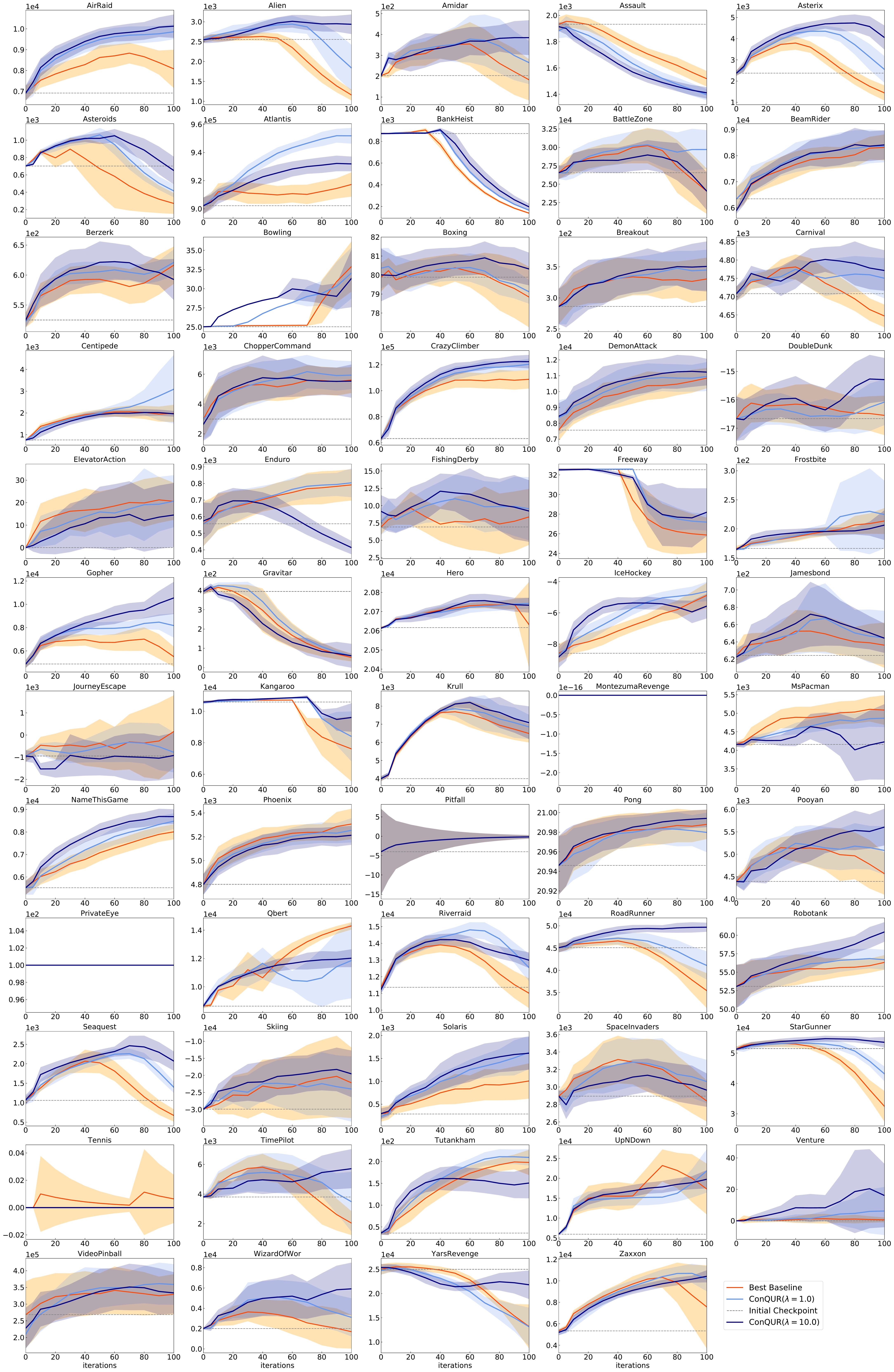}}
\caption{Training curves on 16 nodes with up to 30 no-op starts.}
\label{fig:allgames_allreg}
\end{center}
\end{figure*}

Fig.~\ref{fig:allgames_allreg} shows training curves of \conqur{} with 16 nodes
under different
penalization strengths $\lambda\in \{1, 10\}$. While each game has its own optimal $\lambda$, in general, we found that $\lambda=10$ gave the best performance for most games. Each plotted step of each training curve (including the
baseline) shows the best performing node's policy value as evaluated with full rollouts.
Table~\ref{tab:allnoop} shows the summary of the highest policy values achieved for
all 59 games for \conqur{} and the baseline 
under 16 nodes. 
Both the baseline and \conqur{} improve overall, but \conqur{}'s advantage
over the baseline is amplified. 
These results all use
a splitting factor of $c=4$. 
(We show results with 8 nodes and a splitting factor of 2 below.)

\begin{table*}[t]
\centering
{\small
\begin{tabular}{|l|r|r|r|r|r|} \hline
& \conqur{($\lambda_{\text{best}}$)} (16 nodes) &  Baseline (16 nodes) & Checkpoint \\
\hline
AirRaid & \textbf{10613.01} & 9656.21  & 6916.88 \\ 
Alien  & \textbf{3253.18} & 2713.05 & 2556.64 \\ 
Amidar  & \textbf{555.75}  &446.88 & 203.08 \\ 
Assault  & 2007.81 & \textbf{2019.99} & 1932.55 \\ 
Asterix  & \textbf{5483.41}  & 4649.52 & 2373.44 \\ 
Asteroids  & \textbf{1249.13}  & 1196.56 & 701.96 \\ 
Atlantis  & \textbf{958932.00} & {931416.00} & 902216.00 \\ 
BankHeist  & \textbf{1002.59}  & {965.34} & 872.91 \\ 
BattleZone  &  {31860.30}  & \textbf{32571.80} & 26559.70 \\ 
BeamRider  & 9009.14  & \textbf{9052.38}  & 6344.91 \\ 
Berzerk  & \textbf{671.95}  & 664.69 & 525.06 \\ 
Bowling & {38.36} & \textbf{39.79} & 25.04 \\ 
Boxing  & \textbf{81.37}  & 81.26 & 80.89 \\ 
Breakout  & \textbf{372.31}  & 359.17 & 286.83 \\ 
Carnival  & \textbf{4889.19}  & 4860.66& 4708.14 \\ 
Centipede  &  \textbf{4025.57}   & {2408.23} & 758.21 \\ 
ChopperCommand & \textbf{7818.22} & 6643.07 & 2991.00 \\ 
CrazyClimber  & \textbf{134974.00}  & 119194.00  & 63181.14 \\ 
DemonAttack  & \textbf{11874.80} & 11445.20 & 7564.14 \\ 
DoubleDunk  & \textbf{-14.04} & -15.25  & -16.66 \\ 
ElevatorAction & 24.67  & \textbf{28.67} & 0.00 \\ 
Enduro  & \textbf{879.84} & 835.11 & 556.97 \\ 
FishingDerby  & \textbf{16.28}  & 13.22 & 6.92 \\ 
Freeway  & 32.65  & 32.63 & 32.52 \\ 
Frostbite  & \textbf{289.25}  & 230.29 & 166.44 \\ 
Gopher  & \textbf{11959.20}  & 9084.00 & 4879.02 \\ 
Gravitar & \textbf{489.22}  & 446.64 & 394.46 \\ 
Hero  & \textbf{20827.00} & 20765.70 & 20616.30 \\ 
IceHockey  &  \textbf{-3.15} & {-3.55}  & -8.59 \\ 
Jamesbond  & \textbf{710.78}   & 681.05 & 624.36 \\ 
JourneyEscape & {902.22}  & \textbf{1437.06} & -947.18 \\ 
Kangaroo  & \textbf{11017.65}& 10743.10 & 10584.20 \\ 
Krull  & \textbf{9556.53}   & 9487.49 & 3998.90 \\ 
MontezumaRevenge  & 0.00 & 0.00 & 0.00 \\ 
MsPacman  & {5444.31}  & \textbf{5487.12}  & 4160.50 \\ 
NameThisGame  & \textbf{9104.40} & 8445.43  & 5524.73 \\ 
Phoenix & 5325.33  & \textbf{5430.49} & 4801.18 \\ 
Pitfall  & 0.00& 0.00 & -4.00 \\ 
Pong & 21.00 & 21.00 & 20.95 \\ 
Pooyan & \textbf{5898.46} & 5728.05 & 4393.09 \\ 
PrivateEye  & 100.00  & 100.00 & 100.00 \\ 
Qbert  & 13812.40  & \textbf{15189.00} & 8625.88 \\ 
Riverraid  & \textbf{15895.10} & 15370.10 & 11364.90 \\ 
RoadRunner  & \textbf{50820.40}  & 47481.10 & 45073.25 \\ 
Robotank  & \textbf{62.74} & 57.66 & 53.08 \\ 
Seaquest  & \textbf{3046.34} & 2691.88 & 1060.77 \\ 
Skiing  & \textbf{-13638.80}  & -14908.21 & -29897.07 \\ 
Solaris  & \textbf{1991.33}  & 1202.89 & 285.46 \\ 
SpaceInvaders  & \textbf{3556.10}  & 3520.96 & 2895.30 \\ 
StarGunner  & \textbf{55679.27}  & 55176.90 & 51490.60 \\ 
Tennis  & 0.00  & 0.00 & 0.00 \\ 
TimePilot  & {6698.88}  & \textbf{7327.71} & 3806.54 \\ 
Tutankham  & \textbf{252.51}   & 220.90 & 36.07 \\ 
UpNDown  & {31258.84} & \textbf{34455.20} & 5956.24 \\ 
Venture & \textbf{37.37} & 3.64 & 0.00 \\ 
VideoPinball  & \textbf{423012.59}  & 383105.41 & 268476.04 \\ 
WizardOfWor  & \textbf{8154.73}  & 4782.11  & 2012.24 \\ 
YarsRevenge  & 26188.24  & \textbf{26330.31}  & 25000.36 \\
Zaxxon  & \textbf{11723.20} & {11589.90}  & 5334.44 \\ 
\hline
\end{tabular}
}
\caption{Summary of scores with $\varepsilon$-greedy ($\varepsilon=0.001$) evaluation with up to 30 no-op starts. We ran \conqur{} with 16 nodes and with $\lambda\in \{1, 10\}$. We report the scores of the best  $\lambda_{\text{best}}$ for each game. For most games, $\lambda_{\text{best}}$ is $10$.}
\label{tab:allnoop}
\end{table*}

\newpage
\begin{table*}
\centering
\scalebox{0.97}{
\begin{tabular}{ |p{4cm}| p{7.2cm}|c| }\hline 
\textbf{Hyperparameters}  & \textbf{Description} & \textbf{Value}\\
\hline
Dive levels $d$ to run
&
We run $d$ levels of diving phase after each expansion phase
& 
9\\
\hline
Boltzmann Iteration
&
Every module this number of iteration/level, Q-labels are generated from Boltzmann distribution in order to create diversified node.
& 
5\\
\hline
Online network target network swap frequency
&
Iteration (Frequency) at which the online network parameters swap with the target network
& 
5\\
\hline
Evaluation frequency
&
Iteration (Frequency) at which we perform rollout operation (testing with the environment).
& 
10\\
\hline
Learning Rate
&
Learning rate for the optimizer.
& 
$2.5\times 10^{-6}$\\
\hline
Optimizer
&
Optimizer for training the neural network.
& 
RMSprop\\
\hline
Iteration training data transition size
&
For each iteration, we generate this number of transitions and use it as training data.
& 
10k \\
\hline
Training step frequency
&
For each iteration, we perform (iteration training data transition size / training step frequency) number of gradient updates.
& 
4\\
\hline
Mini-batch size
&
Size of the mini batch data used to train the Q-network.
& 
32\\
\hline
$\varepsilon_{\textrm{train}}$
&
$\varepsilon$-greedy policy for exploration during training.
& 
0.01\\
\hline
$\varepsilon_{\textrm{eval}}$
&
$\varepsilon$-greedy policy for evaluating Q-regressors.
& 
0.001\\
\hline
Training calibration parameter 
&
Calibration to adjust the difference between the nodes from the candidate pool $m$ which didn't selected during both the expansion nor the dive phases. The calibration is performed based on the average difference between the frontier nodes and their parents. We denote this difference as $\bigtriangleup$.
& 
$2.5\bigtriangleup$\\
\hline
Temperature $\tau$
&
Temperature parameter for Boltzmann sampling. Adaptively multiplied or divided by a factor of 1.5 or 4 respectively.
&
1 \\
\hline
Discount factor $\gamma$
&
Discount factor during the training process.
& 
0.99\\
\hline
\end{tabular}

}
\caption{Common Hyperparameters for \conqur{} training and evaluation.}
\label{tab:hps}
\end{table*}

\begin{table*}
\centering
\scalebox{0.97}{
\begin{tabular}{ |p{4cm}| p{7.2cm}|c|c| }\hline 
\textbf{Hyperparameters}  & \textbf{Description} &  \textbf{Value} \\
\hline
Splitting factor $c$ & Number of children created from a parent node & 4  \\
\hline
Candidate pool size $m$ &
Pool of candidate leaf nodes for selection into the dive or expansion phase &
46 \\
\hline
Maximum frontier nodes $F$ &     
Maximum number of child leaf nodes for the dive phase & 
16 \\
\hline  
Top nodes to expand $l$
&
Select the top $l$ nodes from the candidate pool for the expansion phase.
& 4\\
\hline
\end{tabular}
}
\caption{Hyperparameters for \conqur{} (16 nodes) training and evaluation.}
\label{tab:hps_nodes}
\end{table*}


\begin{figure*}[htb]
\begin{center}
    \centerline{\includegraphics[width=0.70\textwidth, scale=0.9]{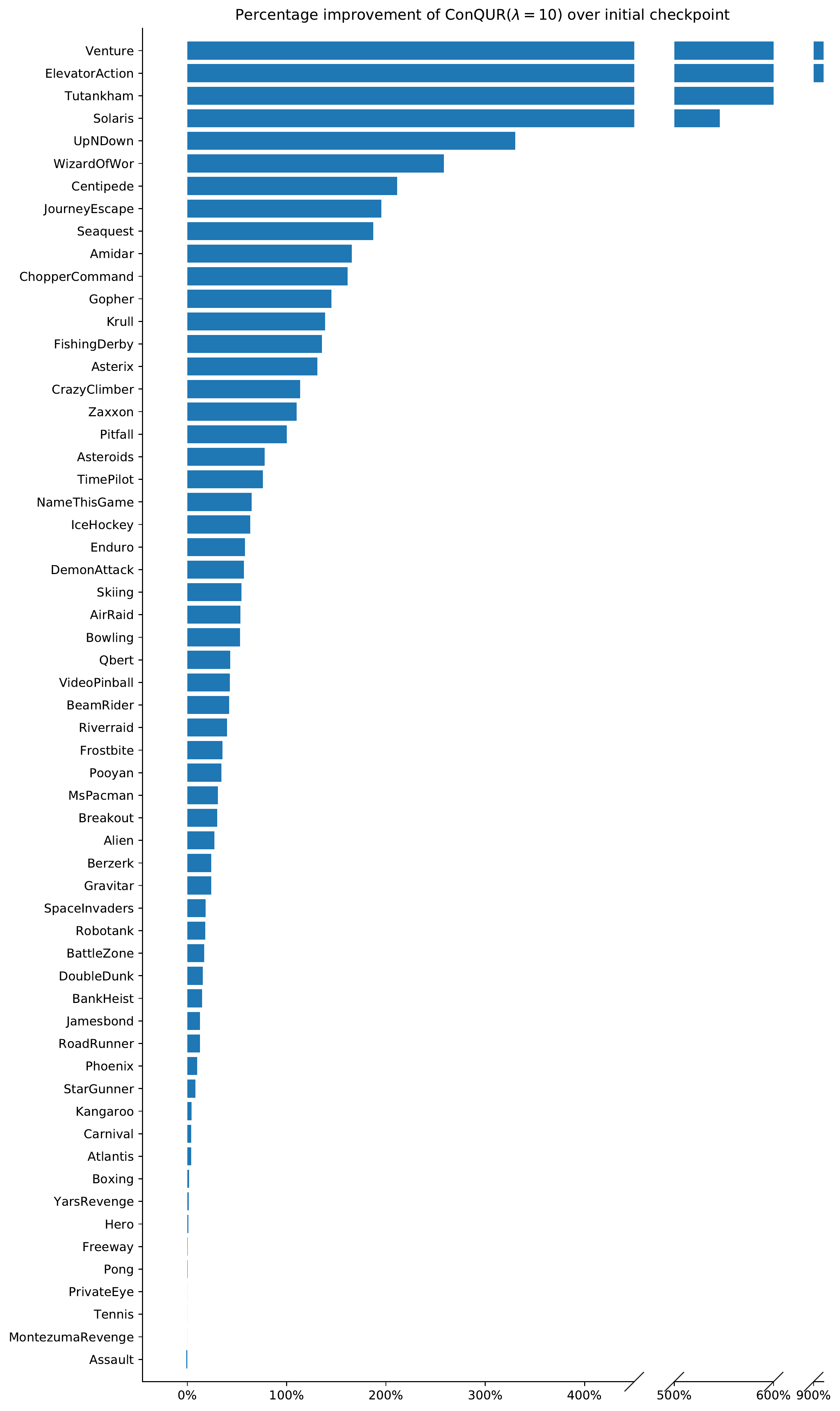}}
\caption{Improvement \conqur{($\lambda=10$)} with 16 nodes achieves over the initial checkpoint Q-network.}
\label{fig:allgames_allreg_initial}
\end{center}
\end{figure*}

\subsection{Additional Results: \conqur{} with 8 Nodes}

As an additional study of \conqur{}, we present results of the running our method using 8 nodes (rather than the 16 used above), and compare it to a multi-DQN baseline that also uses 8 ``nodes'' (i.e., 8 separate DQN runs). We use a splitting factor $c=2$ for $\conqur{}.$ Table~\ref{tab:allnoop_8} shows the average scores for each game using \conqur{} and the baseline with 8 nodes. Unsurprisingly, \conqur{} with 8 nodes does not perform as well as \conqur{} with 16 nodes; but as in the 16-node case, \conqur{} outperforms the baseline when each uses 8 nodes. More importantly, the average improvement of $24.5\%$ for \conqur{} with 16 nodes over the corresponding baseline \emph{exceeds} the $19.6\%$ improvement of \conqur{} in the 8-node case. This is a strong indication that increasing the number of nodes increases the performance gap \emph{relative} to the corresponding multi-DQN baseline; this implies that a good search heuristic is critical to effectively navigate the search space (as compared to randomly selected nodes) with a greater number of candidate hypotheses.\footnote{Average score improvements exclude games where the baseline score is zero.}

\begin{table*}[t]
\centering
{\small
\begin{tabular}{|l|r|r|r|r|r|} \hline
& \conqur{($\lambda_{\text{best}}$)} (8 nodes) &  Baseline (8 nodes) & Checkpoint \\
\hline
AirRaid & \textbf{10647.80} & 9050.86  & 6885.72 \\ 
Alien  & \textbf{3341.36} & 3207.5.05 & 2556.64 \\ 
Amidar  & \textbf{577.45}  &573.55 & 202.74 \\ 
Assault  & 1892.02 & \textbf{1976.80} & 1873.05 \\ 
Asterix  & \textbf{5026.24}  & 4935.21 & 2373.44 \\ 
Asteroids  & \textbf{1194.90}  & 1170.11 & 704.38 \\ 
Atlantis  & \textbf{949012.00} & {932668.00} & 902216.00 \\ 
BankHeist  & {909.61}  & \textbf{924.75} & 871.91 \\ 
BattleZone  &  \textbf{32139.90}  & {30983.10} & 26558.70 \\ 
BeamRider  & \textbf{8613.98}  & {8109.63}  & 6397.49 \\ 
Berzerk  & \textbf{659.64}  & 634.83 & 524.76 \\ 
Bowling & \textbf{30.07} & \textbf{25.29} & 25.04 \\ 
Boxing  & \textbf{81.78}  & 81.48 & 80.29 \\ 
Breakout  & {350.11}  & \textbf{362.98} & 286.14 \\ 
Carnival  & \textbf{4862.30}  & 4800.83 & 4708.23 \\ 
Centipede  &  \textbf{2747.89}   & {2608.78} & 757.51 \\ 
ChopperCommand & \textbf{7188.25} & 6737.21 & 2641.71 \\ 
CrazyClimber  & \textbf{131675.00}  & 122424.00  & 63181.11 \\ 
DemonAttack  & \textbf{11346.20} & 10947.90 & 8022.08 \\ 
DoubleDunk  & \textbf{-13.57} & -15.35  & -16.66 \\ 
ElevatorAction & \textbf{28.00}  & \textbf{21.33} & 0.00 \\ 
Enduro  & \textbf{849.07} & 811.58 & 556.56 \\ 
FishingDerby  & \textbf{13.34}  & 11.56 & 7.15 \\ 
Freeway  & 32.60  & 32.60 & 32.52 \\ 
Frostbite  & \textbf{296.57}  & 220.81 & 165.01 \\ 
Gopher  & \textbf{9999.61}  & 8013.34 & 4879.13 \\ 
Gravitar & {475.03}  & \textbf{480.64} & 394.46 \\ 
Hero  & \textbf{20803.60} & 20774.80 & 20598.40 \\ 
IceHockey  &  \textbf{-3.23} & {-4.78}  & -8.63 \\ 
Jamesbond  & {664.98}   & \textbf{669.54} & 626.53 \\ 
JourneyEscape & {-462.64}  & \textbf{391.44} & -947.18 \\ 
Kangaroo  & \textbf{10974.00}& 10733.60 & 10584.20 \\ 
Krull  & {9503.62}   & \textbf{9538.22} & 4039.78 \\ 
MontezumaRevenge  & \textbf{1.46} & 0.00 & 0.00 \\ 
MsPacman  & {5066.17}  & \textbf{5227.84}  & 4160.50 \\ 
NameThisGame  & \textbf{9181.30} & 8410.29  & 5529.50 \\ 
Phoenix & \textbf{5307.46}  & {5227.84} & 4801.18 \\ 
Pitfall  & 0.00& 0.00 & -4.00 \\ 
Pong & 21.00 & 20.99 & 20.95 \\ 
Pooyan & \textbf{5778.99} & 5184.14 & 4393.09 \\ 
PrivateEye  & 100.00  & 100.00 & 100.00 \\ 
Qbert  & 11953.40  & \textbf{13965.80} & 8625.88 \\ 
Riverraid  & \textbf{15614.40} & 14812.40 & 11253.30 \\ 
RoadRunner  & \textbf{49864.80}  & 46302.20 & 45073.25 \\ 
Robotank  & \textbf{61.92} & 56.90 & 53.08 \\ 
Seaquest  & \textbf{2647.82} & 2560.61 & 1034.83 \\ 
Skiing  & \textbf{-14058.90}  & -14079.80 & -29896.80 \\ 
Solaris  & \textbf{1956.24}  & 1182.59 & 291.70 \\ 
SpaceInvaders  & \textbf{3436.16}  & 3292.68 & 2895.30 \\ 
StarGunner  & \textbf{55479.00}  & 54207.30 & 51419.60 \\ 
Tennis  & 0.00  & 0.00 & 0.00 \\ 
TimePilot  & {6717.62}  & \textbf{6799.19} & 3806.22 \\ 
Tutankham  & \textbf{242.03}   & 229.23 & 36.00 \\ 
UpNDown  & \textbf{22544.60} & {23331.20} & 5956.21 \\ 
Venture & \textbf{15.41} & 1.50 & 0.00 \\ 
VideoPinball  & {382110.59}  & \textbf{390540.41} & 209620.0 \\ 
WizardOfWor  & \textbf{5750.05}  & 3632.17  & 2011.77 \\ 
YarsRevenge  & \textbf{25631.10}  & {25396.70}  & 25319.20 \\
Zaxxon  & \textbf{10751.80} & {11156.20}  & 5186.01\\
\hline
\end{tabular}
}
\caption{Summary of scores with $\varepsilon$-greedy ($\varepsilon=0.001$) evaluation with up to 30 no-op starts. As a side study, we ran \conqur{} with 8 nodes and with $\lambda\in \{1, 10\}$. We report the scores of the best  $\lambda_{\text{best}}$ for each game. For most games, $\lambda_{\text{best}}$ is $10$. The 8 nodes configuration follows the same as in Table~\ref{tab:hps}, except that $c=2, m=38, F=8, l=2.$}
\label{tab:allnoop_8}
\end{table*}



\end{document}